%% file: paper.tex
\documentclass[10pt,journal]{IEEEtran}

\usepackage[utf8]{inputenc}
\usepackage[table]{xcolor}
\usepackage{rotating}

\relax

\input{include}
\input{my_definitions.tex}

\usepackage{array}
\usepackage{footnote}
\makesavenoteenv{tabular}
\makesavenoteenv{table}

\newcommand{\transpose}{^\top}

\newcommand{\wbigcup}{\mathop{{\bigcup}}\displaylimits}
\usepackage{booktabs}       % professional-quality tables
\title{Edge Dithering for Robust Adaptive Graph Convolutional Networks}
\author{Vassilis N. Ioannidis, and
Georgios B. Giannakis
\thanks{
The work in this paper has been supported by the Doctoral Dissertation Fellowship of the Univ. of Minnesota, the USA NSF grants 171141,  1500713, and  1442686.}
\\ECE Dept. and Digital Tech. Center, Univ. of Minnesota, Mpls, MN 55455, USA\\
E-mails: $\{$ioann006, georgios$\}$ @umn.edu
}

\begin{document}
	
	\maketitle
	\begin{abstract}
		Graph convolutional networks (GCNs) are vulnerable to perturbations of the graph structure that are either random, or, adversarially designed. The perturbed  links modify the graph neighborhoods, which critically affects the performance of GCNs in semi-supervised learning (SSL) tasks.  Aiming at robustifying GCNs  conditioned on the perturbed graph, the present paper generates multiple auxiliary graphs, each having 
		its binary $0-1$ edge weights flip values with probabilities designed to enhance robustness. The resultant edge-dithered auxiliary  graphs are leveraged by an adaptive (A)GCN that performs SSL. Robustness is enabled through learnable graph-combining weights along with suitable regularizers. Relative to GCN, the novel AGCN achieves markedly improved performance in tests with noisy inputs, graph perturbations, and  state-of-the-art adversarial attacks. Further experiments with protein interaction networks showcase the competitive performance of AGCN for SSL over multiple graphs.
	\end{abstract}
	
\section{Introduction}
A task of major importance at the cross-roads of machine learning and network science is semi-supervised learning (SSL) over graphs. SSL aims at predicting nodal labels given: i) the graph connections; ii) feature vectors at all nodes; and iii) labels only at a subset of nodes.  This partial label availability may be attributed to privacy concerns (e.g., with medical data); energy considerations  (e.g., with wireless sensor networks); or unrated items (e.g., with recommender systems).
	
Standard SSL schemes typically assume that the available labels and graph connections have certain properties such as smoothness, which asserts that connected nodes have similar attributes~\cite{smola2003kernels}. In various scenarios however, robustness issues arise.  Powerful adversaries manipulate nodal attributes and connections to bias learning, and promote their malicious goals~\cite{zugner18adv}. Further, human annotators or noisy data introduce errors during the graph construction that leads to perturbed edge weights~\cite{akoglu2015graph}. Adversarially perturbed or simply anomalous graph data may degrade the performance of SSL {algorithms} with severe consequences. The recent era of misinformation and ``fake'' news calls for robust machine learning algorithms for network science~\cite{goodfellow2014explaining,aggarwal2015outlier,yan2016robust}. In this context, a novel robust GCN framework is introduced here that utilizes edge-dithered auxiliary graphs, which are combined using learnable weights.

\subsection{Related work}
Graph-based SSL methods typically assume that the true labels are ``smooth'' with respect to the underlying 
network structure, which naturally motivates leveraging the topology of the network to propagate the labels and increase classification performance. Graph-induced smoothness may be captured by kernels on graphs~\cite{belkin2006manifold,smola2003kernels}; or  Gaussian random fields \cite{zhu2003semi}. Graph convolutional networks (GCN)s incorporate the graph structure to achieve state-of-the-art results in SSL tasks \cite{kipf2016semi,bronstein2017geometric,velivckovic2017graph,xu2018powerful}. 
	
With the success of GCNs on graph learning tasks granted, recent results indicate that perturbations of the graph topology or nodal features can severely deteriorate their classification performance \cite{zugner18adv,xu2019topology,dai2018adversarial}. Structural attacks target a subset of nodes and modify their links to promote miss-classification of targeted nodes~\cite{wu19adv}.  The designed graph perturbations are ``unnoticeable'', which is feasible so long as the degree distribution of the perturbed graphs are similar to the initial distribution~\cite{zugner18adv}.  GCNs learn nodal representations by extracting information within local neighborhoods. Adversaries poison the learned features by perturbing the node's neighborhood. Hence, the vulnerability of GCNs challenges their deployment to critical applications dealing with security or healthcare, where robust learning is of major importance. Defending against adversaries may unleash the potential of GCNs and broaden the scope of machine learning applications altogether. Recent works robustify GCNs against structural perturbations by utilizing the nodal features~\cite{wu19adv,zhu2019robust}. Gaussian regularizers are employed in \cite{zhu2019robust} to protect the network from adversarial attacks. Jaccard similarity among features is utilized in \cite{wu19adv} to prune perturbed edges.  However, these methods are challenged in the absence of nodal feature vectors.
	
\subsection{Contributions} 
The present paper develops a framework for robust deep learning over perturbed graphs. Specifically, 
the contribution of this work is threefold.
\begin{itemize}
\item[\textbf{C1}.]  Given the perturbed unweighted graph and aiming at robust SSL, multiple auxiliary graphs are drawn by dithering (adding or removing) edges with probabilities selected to boost robustness. The novel edge-dithering (ED) approach reconstructs the original neighborhood structure with high probability (whp) as the number of sampled graphs increases. ED can be applied even in the absence of nodal features.
\item[\textbf{C2}.] A weighted combination of the auxiliary ED graphs is employed across GCN layers. Per layer weights are adapted to promote those ED graphs that maximally avoid the adversarially perturbed edges. Further, a residual feed of the data is utilized to facilitate diffusion of the features across the graph. Robust graph-based regularizers are also included to prevent overfitting, and further account for the underlying graph topology. 
\item[\textbf{C3}.] Numerical tests involving noisy features, noisy edge weights, and random as well as adversarial edge perturbations showcase the merits of the novel approach. The proposed ED-AGCN achieves also competitive results in predicting protein functions over multiple graphs. 
\end{itemize}
	
\section{Modeling and problem formulation}
Consider a graph $\mathcal{G}\define(\mathcal{V},\mathbf{A})$ of $N$ nodes, with $\mathcal{V}\define\{v_1,\ldots,v_N\}$ denoting the vertex set, and $\mathbf{A}$ the $N\times N$ adjacency matrix capturing edge connectivity through $A_{n,n'}$ that is $1$ if an edge connects $v_n$ and $v_n'$, and 0 otherwise. The neighborhood of  $v_n$ is 
\begin{align}
	\label{eq:neighborhoodor}
	\mathcal{N}_n\define\{n':A_{n,n'}\ne0,~~ v_n'\in\mathcal{V}\}.
	\end{align}
The perturbed graph is $\bar{\mathcal{G}}\define(\mathcal{V},\bar{\mathbf{A}})$ with corresponding adjacency $\bar{\mathbf{A}}\define\mathbf{A}+\check{\mathbf{A}}$ having entries perturbed by  
	\begin{align}
	\check{A}_{n,n'}=\begin{cases}
	1,& \text{if}~A_{n,n'}=0~\text{and}~\bar{A}_{n,n'}=1 \\
	-1,& \text{if}~A_{n,n'}=1~\text{and}~\bar{A}_{n,n'}=0 \\
	0,& \text{otherwise}
	\end{cases}
	\end{align}
where $+1$ corresponds to edge insertion, $-1$ to edge deletion, and $0$ to no perturbation. Evidently, these links may drastically degrade the performance of SSL methods since the neighborhood is either adversarially or randomly modified~\cite{zugner18adv}. The adversarial attacks aim at \emph{unnoticeable} changes, a constraint that limits the number of perturbations. Hence, the number of perturbed links (nonzero elements in $\check{\mathbf{A}}$) is small relative to the original number of edges in $\mathcal{G}$.  
	
Associated with the $n$-th node can be an $F\times 1$ feature vector $\mathbf{x}_{n}$. These vectors are collected in the $N\times F$ feature matrix  $\mathbf{X}\define[\mathbf{x}_{1}\transpose,\ldots,\mathbf{x}_{N}\transpose]\transpose$, where  $X_{nf}$ may denote, for example, the salary of the  $n$-th individual in the LinkedIn social network. Let also  $y_n\in\{0,1,\ldots,K-1\}$ denote the label of node $n$, which may represent, for example, the education level of a person.  The $N\times K$ matrix  $\mathbf{Y}$ is the ``one-hot'' representation of the nodal labels belonging to $K$ classes, that is, if $y_n=k$ then $Y_{n,k}=1$ and $Y_{n,k'}=0, \forall k'\ne k$.
	
\noindent
\textbf{Goal}. Given the perturbed topology $\bar{\mathbf{A}}$, the features in $\mathbf{X}$, and labels only at a subset $\mathcal{L}$ of nodes $\{y_{n}\}_{n\in\mathcal{L}}$ with $\mathcal{L} \subset\mathcal{V}$, the goal of this paper is to design robust GCN architectures that are minimally affected by the perturbed edges.

\section{Edge dithering}
The ever-expanding interconnection of social, email, and media service platforms presents an opportunity for adversaries manipulating networked data to launch malicious attacks~\cite{goodfellow2014explaining,aggarwal2015outlier,zugner18adv}. Perturbed  edges modify the graph neighborhoods, which leads to significant degradation in the performance of GCNs. Aiming to restore a node's initial graph neighborhood an edge-dithering (ED) module is developed in this section, where auxiliary graphs are created with probabilities designed to enhance robustness.  Dithering in visual and audio applications, refers to intentional injection of noise so that the quantization error is converted to random noise, which can be easily handled~\cite{ulichney1988dithering}. 
	
Permeating the benefits of dithering towards robustify GCNs, we generate ED graphs $\{\mathcal{G}_{i}\}_{i=1}^{I}$, where $\mathcal{G}_{i}\define(\mathcal{V},\mathbf{A}_{i})$. Each auxiliary graph $\mathcal{G}_{i}$ is a dithered version of the perturbed graph $\bar{\mathcal{G}}$, where the edges in $\mathbf{A}_{i}$ are selected in a probabilistic fashion as follows
\begin{align}
	\label{eq:samplegraph}
	A_{n,n',i}=\left\{
	\begin{array}{ll}
	1& \text{wp.}~~~q_1^{ \delta(\bar{A}_{n,n'}=1)}{(1-q_2)}^{  \delta(\bar{A}_{n,n'}=0)}\\
	0&\text{wp.}~~~q_2^{ \delta(\bar{A}_{n,n'}=0)}{(1-q_1)}^{  \delta(\bar{A}_{n,n'}=1)}
	\end{array} 
	\right. 
	\end{align}
where $\delta(\cdot)$ is the indicator function, $q_1:={\rm Pr}(A_{n,n',i}=1|\bar{A}_{n,n'}=1)$ and 
$q_2:={\rm Pr}(A_{n,n',i}=0|\bar{A}_{n,n'}=0)$. If $n$ and $n'$ are connected in $\bar{\mathcal{G}}$, the edge connecting $n$ with $n'$ is deleted with probability $1-q_1$. Otherwise, if $n$ and $n'$ are not connected in $\bar{\mathcal{G}}$ i.e. $(\bar{A}_{n,n'}=0)$, an edge between $n$ and $n'$ is inserted with probability $1-q_2$. 

Hence, the $i$th ED graph neighborhood of $v_n$ is 
\begin{align}
	\label{eq:neighborhood}
	\mathcal{N}_n^{(i)}\define\{n':A_{n,n',i}\ne0,~~ v_n'\in\mathcal{V}\}.
	\end{align}
The ED graphs give raise to different neighborhoods $\mathcal{N}_n^{(i)}$, and the role of ED is to ensure that the unperturbed neighborhood of each node will be present with high probability (whp) in at least one of the $I$ graphs. The ensuing remarks assert that this will happen whp as $I$ increases. 
\begin{myremark}
With high probability, there exists $\mathcal{G}_i$ such that a perturbed edge will be restored to its initial value. This means that there exists an ED graph $i$ such that $A_{n,n',i}=A_{n,n'}$. Since, each $\mathcal{G}_i$ is independently drawn, it holds that 
		\begin{align}
		{\rm Pr}\bigg(\wbigcup_{i=1}^I(A_{n,n',i}=0)\Big|\bar{A}_{n,n'}=1,A_{n,n'}=0\bigg)=1-q_1^I\nonumber\\
		\nonumber{\rm Pr}\bigg(\wbigcup_{i=1}^I(A_{n,n',i}=1)\Big|\bar{A}_{n,n'}=0,A_{n,n'}=1\bigg)=1-q_2^I
		\end{align}
\end{myremark}
	
\begin{myremark}
Whp there exists $\mathcal{G}_i$ which will recover the original neighborhood structure of a node, i.e. $\mathcal{N}_n^{(i)}=\mathcal{N}_n$. The proof of this remark is included in the Appendix.
\end{myremark}
	
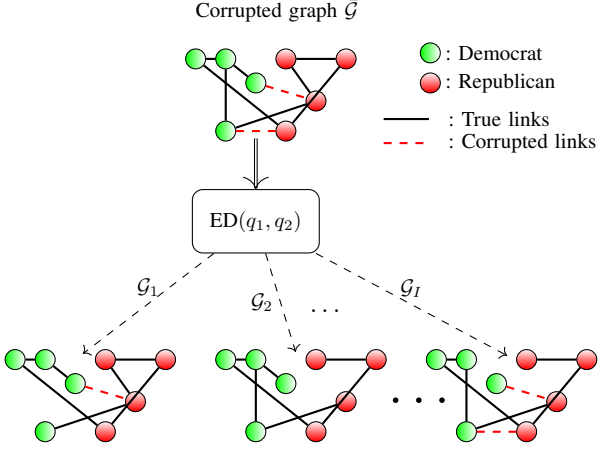
\begin{figure}
		\centering
		\input{figs/sampledGraph.tex}
		\caption{ED in operation on a perturbed social network among voters. Black solid edges are the true links and dashed red edges represent adversarially perturbed links.}
		\label{fig:sampled}
\end{figure}

The high probability claims asserted in Remarks 1 and 2 hold as $I$ increases. Nevertheless, experiments with adversarial attacks demonstrate that even with a small $I$ the use of ED significantly boosts classification performance. The operation of the ED module is detailed in Fig.~\ref{fig:sampled}. Note that the proposed ED does not require availability of nodal feature vectors. The generated graphs have to be processed by a dedicated architecture that promotes the learned features from unperturbed nodal neighborhoods. 
	
\section{Adaptive GCN with edge dithering}
The ED module generates $\{\mathcal{G}_i\}_{i=1}^{I}$, which along with the perturbed graph $\bar{\mathcal{G}}$ will be judiciously combined to obtain a robust learning architecture. Typically, deep or shallow learning over graphs considers that the relation among the nodal variables is represented by a single graph. This may be inadequate in several contemporary applications, where nodes may engage on multiple relations~\cite{kivela2014multilayer}, motivating the generalization of SSL approaches for \emph{single} graphs to \emph{multiple} graphs\footnote{{Many works refer to these as multi-layer graphs~\cite{kivela2014multilayer}.}}. In social networks for example, each graph may capture a specific form of social interaction, such as friendship, family bonds, or coworker-ties \cite{wasserman1994social}. Aiming at a weighted combination of the auxiliary ED graphs, this section develops a novel GCN that adapts to multiple relations and enhances robustness.   
	
Deep learning architectures typically process the input information using $\nbrlayers$ hidden layers. Each layer implements a conveniently parametrized linear transformation, a scalar nonlinear transformation, and oftentimes a dimensionality reduction (pooling) operator. Through nonlinear mappings of linearly combined local features the idea is to progressively extract useful information~\cite{goodfellow2016deep}. GCNs tailor these operations to the graph that supports the data \cite{bronstein2017geometric,kipf2016semi,cao2016deep,li2018adaptive}. Next, our AGCN architecture and training are presented.

\subsection{Per layer operation}
Consider a hidden layer (say the $l$th one), whose output is the  $N\times I\times P^{(l)}$ tensor $\underline{\check{\mathbf {Z}}}^{(l)}$ that holds the $ P^{(l)}\times 1$  feature vectors 	$\check{\mathbf{z}}_{n,i}^{(l)}, \forall n,i$, with $ P^{(l)}$ being the number of output features at $ l$. Similarly, let $\underline{\check{\mathbf {Z}}}^{(l-1)}$ represent the input to this $l$th layer.

The mapping from $\underline{\check{\mathbf {Z}}}^{(l-1)}$ to $\underline{\check{\mathbf {Z}}}^{(l)}$ can be split into two steps. A linear one designed to map the tensor $\underline{\check{\mathbf {Z}}}^{(l-1)}$ to the tensor $\underline{\mathbf{Z}}^{(l)}$.  The latter is then processed elementwise to obtain   $\underline{\check{Z}}_{i,n,p}^{(l)}\define \sigma{(\underline{{Z}}_{i,n,p}^{(l)})}$. A common choice for $ \sigma{(\cdot)}$ is the rectified linear unit (ReLU), for which $\sigma{(c)}=\text{max}(0,c)$.

Of critical importance is the design of the linear map from $\underline{\check{\mathbf {Z}}}^{(l-1)}$ to $\underline{\mathbf{Z}}^{(l)}$ that is tailored to our ED-based setup. Convolutional (C)NNs typically consider a small number of trainable weights, and then generate the linear output by convolving the input with these weights~\cite{goodfellow2016deep}. The convolution combines values of close-by inputs (consecutive time instants, or neighboring pixels), and thus extracts information of local neighborhoods. 

GCNs generalize CNNs to operate on graph data by replacing the convolution with a graph filter whose parameters are also learned~\cite{bronstein2017geometric,kipf2016semi}. This preserves locality, reduces the degrees of freedom of the map, and leverages the graph structure.

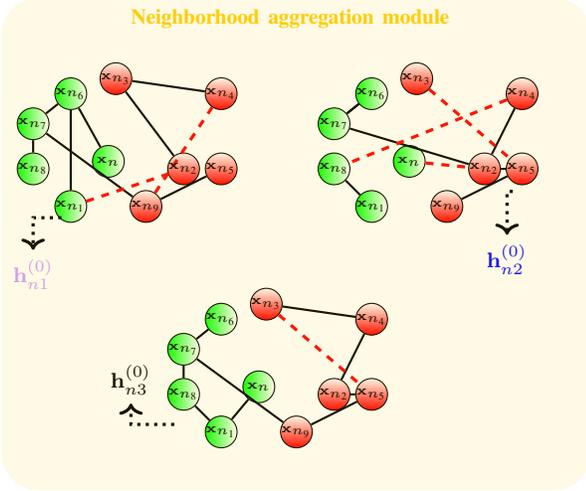
\begin{figure}
		\centering
		\input{figs/neighboragrmodule.tex}
		\caption{The neighborhood aggregation module (NAM) combines features $\{\mathbf{x}_n\}$ of the multiple ED graphs. For this example, the 2-hop neighborhood is considered ($K=2$). 
		Different graphs define different local neighborhoods. }
		\label{fig:neiggrmodule}
\end{figure}

\noindent\textbf{Neighborhood aggregation module}. First, a neighborhood aggregation module is considered that combines linearly the nodal features within a {graph} neighborhood. Since the neighborhood depends on the particular ED graph \eqref{eq:neighborhood}, the combined $n$th feature of the $i$-th graph is
\begin{align}
\mathbf{h}_{n,i}^{(l)}
	\define
	\sum_{n'\in\mathcal{N}_n^{(i)}} A_{n,n',i}
	\check{\mathbf{z}}_{n',i}^{(l-1)}.
	\label{eq:sem}
\end{align}
While the entries of $\mathbf{h}_{n,i}^{(l)}$ depend only on the one-hop neighbors of $n$ (one-hop diffusion), successive application of this operation will increase the diffusion range, spreading the information across the network. Generalizing to neighborhoods with larger diameter, consider the $k$th power of the adjacency matrix $\mathbf{A}^k$. Indeed, the vector $\mathbf{A}^k\mathbf{x}$ holds the linear combinations of the values of $\mathbf{x}$ in the $k$-hop neighborhood~\cite{kipf2016semi}. After defining the matrices $\mathbf{A}_i^k\define\mathbf{A}_i^{(k)}$ for $k=1,\ldots,K,~i=1,\ldots,I$, consider the following parametrized mapping
\begin{align}
	\mathbf{h}_{n,i}^{(l)}
	\define\sum_{k=1}^K
	\sum_{n'\in\mathcal{N}_n^{(i)}} c^{(k)}_iA_{n,n',i}^{(k)}
	\check{\mathbf{z}}_{n',i}^{(l-1)},~~ \forall n,i
	\label{eq:gf}
\end{align}
where the learnable coefficients $\{c^{(k)}_i\}_{k=1}^K$ weight the effect of the corresponding $k$-th hop neighbors for relation $i$. At the $ l$-th layer, the coefficients $\{\{c^{(k)}_i\}_{k=1}^K\}_{i=1}^I$ are collected in the $K\times I$ matrix $\mathbf{C}^{(l)}$. The proposed map in \eqref{eq:gf} aggregates the diffused features in the $K$-hop neighborhoods per $i$; see also Fig. \ref{fig:neiggrmodule}. 

\noindent\textbf{Graph adaptive module}.
The extracted feature vector $\mathbf{h}_{n,i}^{(l)}$ captures the diffused features per ED graph $i$. Aiming at robustness, the learning algorithm should promote features originating from non-perturbed graph neighborhoods. Towards this end, a graph adaptive module is developed that combines $\mathbf{h}_{n,i'}^{(l)}$ across $i'$ as 
\begin{align}\label{eq:graphadaptive}
	\mathbf{g}_{n,i}^{(l)}\define&\sum_{i'=1}^I R _{i,i',n}^{(l)}{\mathbf{h} _{n,i'}^{(l)}}
\end{align}
where $ R _{i,i',n}^{(l)}$ mixes features across ED graphs. A key contribution of this paper is viewing $\{ R _{i,i',n}^{(l)}\}_{i,i',n}$ as training parameters, which allows AGCN to learn how to combine the different relations encoded by the ED graphs. This characteristic enables the novel AGCN to navigate through the ED graphs $\{\mathcal{G}_{i}\}_{i=1}^{I}$, and assign larger weights to features originating from non-perturbed neighborhoods. 

The graph adaptive module in \eqref{eq:graphadaptive} allows for different $ R_{i,i',n}$ per $n$. Considering the same $ R$ for each $n$, that is $ R_{i,i',n}^{(l)}= R_{i,i'}^{(l)}$, results in a design with less parameters at the expense of reduced flexibility. On the other hand, the flexible design in \eqref{eq:graphadaptive} allows large weights $ R_{i,i',n}$ for neighborhoods without corrupted edges, even if $\mathcal{G}_i$ is perturbed.
	
\noindent\textbf{Feature aggregation module}.
Next, the graph adaptive features $\mathbf{g}_{n,i}^{(l)}$ are mixed using learnable vectors	$\mathbf{w}_{n,i,p}^{(l)}$ to obtain 
\begin{align}\label{eq:linconv}\underline{{Z}}_{i,n,p}^{(l)}\define&\mathbf{g}_{n,i}
	\transpose\mathbf{w}^{(l)}_{n,i,p},\\&i=1,\ldots,I,~{n}=1,\ldots,N,~{p}=1,\ldots, { P}^{(l)}\;.
	\nonumber
\end{align}
%where  the  $ P^{(l-1)}\times 1$ vector $\mathbf{w}_{n,i,p}^{(l)}$ mixes the features.

The $ P^{(l-1)}\times N\times I\times P^{(l)}$ tensor $\underline{\mathbf{W}}^{(l)}$ collects the feature mixing weights $\{\mathbf{w}_{n,i,p}^{(l)}\}$, while the $I\times I\times P^{(l)}$ tensor $\underline{\mathbf{R}}^{(l)}$ collects the graph mixing weights $\{ R_{i,i',n}^{(l)}\}$. Upon collecting all the scalars $\{\underline{{Z}}_{i,n,p}^{(l)}\}$ in  $\underline{\mathbf{Z}}^{(l)}$ \eqref{eq:gf}-\eqref{eq:linconv} reduce to 
\begin{align}
	\label{eq:lingnn}
	\underline{\mathbf{Z}}^{(l)}&\define
	\layfunc(\underline{\check{\mathbf {Z}}}^{(l-1)};
	\bm{\theta}_z^{(l)})\\
	\label{eq:param}
	\bm{\theta}_z^{(l)}&\define[\text{vec}(\underline{\mathbf{W}}^{(l)});\text{vec}(\underline{\mathbf{R}}^{(l)})
	;\text{vec}(\mathbf{C}^{(l)})]\transpose.
\end{align}
	
\noindent\textbf{Residual GCN layer}.
Concatenating $L$ GCN layers diffuses the input $\mathbf{X}$ across the $L$-hop graph neighborhood, cf.~\eqref{eq:sem}. However, the exact size of the relevant neighborhood is not always known a priori. To endow our architecture with increased flexibility,  a residual GCN layer is introduced that inputs $\mathbf{X}$ at each $ l$, and thus captures multiple types of diffusion\footnote{This is also known as a skip connection~\cite{he2016deep}}. As a result, the linear operation in \eqref{eq:lingnn} is replaced by the residual linear tensor mapping 
\begin{align}
	\underline{\mathbf{Z}}^{(l)}\define
	\layfunc(\underline{\check{\mathbf {Z}}}^{(l-1)};
\bm{\theta}_z^{(l)})+
	\layfunc(\datatensor;
	\bm{\theta}_x^{(l)})
	\label{eq:residuallayer}
\end{align}
where $\bm{\theta}_x^{(l)}$ encodes trainable parameters, cf. \eqref{eq:param}.  When viewed as a map from  $\datatensor$ to $\underline{\mathbf{Z}}^{(l)}$, the operator in  \eqref{eq:residuallayer} implements a broader class of graph diffusions than the one in \eqref{eq:lingnn}. If $l=3$ and $K=1$ for example, the first summand in \eqref{eq:residuallayer} is a one-hop diffusion of the input that corresponds to a two-hop (nonlinear) diffused version of $\mathbf{X}$, while the second summand diffuses $\mathbf{X}$ in one-hop. At a more intuitive level, the second summand also guarantees that the impact of $\mathbf{X}$ in the output does not vanish as the number of layers grows. 

The output of our graph architecture is 
\begin{align}\label{eq:output}
	\hat{\mathbf{Y}} = g(\underline{\check{\mathbf {Z}}}
	^{(L)};\bm{\theta}_g)
\end{align}
where $g(\cdot)$ is the normalized exponential function (softmax), $\hat{\mathbf{Y}}$ is an $N\times K$ matrix, $\hat{{Y}}_{n,k}$ represents the probability that $y_n=k$, and $\bm{\theta}_g$ are trainable parameters. %For notational convenience, the global mapping from $\mathbf{X}$ to $\hat{\mathbf{Y}}$  is denoted as 
%\begin{align}
%	\hat{\mathbf{Y}}\define\mathcal{F}(\mathbf{X};\{\bm{\theta}_z^{(l)}\},\{\bm{\theta}_x^{(l)}\},\bm{\theta}_g),
%\end{align} 
%and represented in the block diagram depicted in.
	
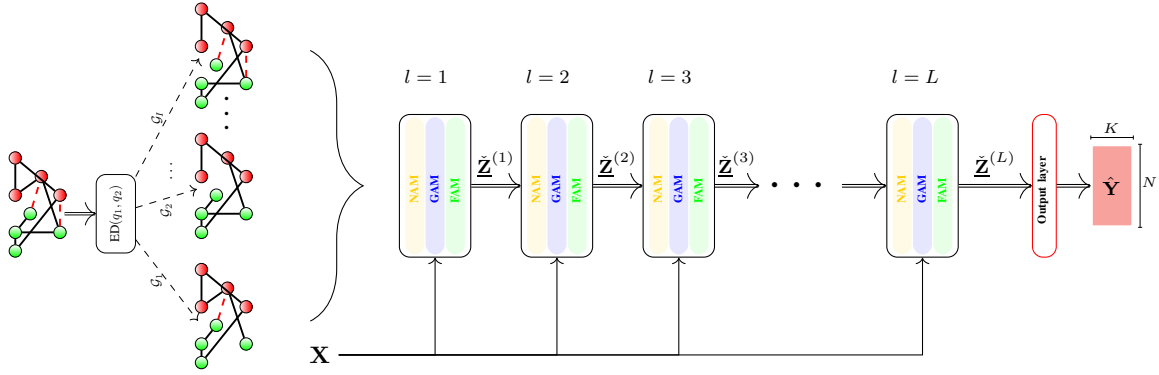
\begin{figure*}
		\centering
		\input{figs/grnn.tex}
		\caption{Given the perturbed graph, the ED module generates the auxiliary graphs, which along with $\mathbf{X}$ are processed by the AGCN.  Each AGCN layer includes a neighborhood aggregation module (NAM), a graph adaptive module (GAM), and a feature aggregation module (FAM). Notice the skip connections that input $\mathbf{X}$ to each layer.}
		\label{fig:grnn}
\end{figure*}
	
\subsection{Training with graph-smooth regularizers}
The AGCN weights are estimated by minimizing the discrepancy between estimated labels and the given ones as
\begin{align}\label{eq:trainobj}    \min_{\{\bm{\theta}_z^{(l)}\},\{\bm{\theta}_x^{(l)}\},\bm{\theta}_g}&
	\mathcal{L}_{tr} (\hat{\mathbf{Y}},\mathbf{Y}) +\regpargraphsmooth\sum_{i=1}^I\text{Tr}(\hat{\mathbf{Y}}\transpose\mathbf{A}_i\hat{\mathbf{Y}})
	\nonumber\\
	+&\regparsmooth\regfun(\{\bm{\theta}_z^{(l)}\},\{\bm{\theta}_x^{(l)}\})
	+\regparsparse \sum_{ l=1}^\nbrlayers\|\underline{\mathbf{R}}^{(l)}\|_1%\\
	%\text{s.t.}&~~
	%\hat{\mathbf{Y}}=\mathcal{F}(
	%\mathbf{X};\{\bm{\theta}_z^{(l)}\},\{\bm{\theta}_x^{(l)}\},\bm{\theta}_g). 
\end{align}
where $\mathcal{L}_{tr} (\hat{\mathbf{Y}},\mathbf{Y})\define-\sum_{n\in\mathcal{L}}\sum_{k=1}^K Y_{n k}\ln{\hat{{Y}}_{n k}}$ is the cross-entropy loss function over the labeled examples.  

The first regularizer in \eqref{eq:trainobj} promotes smooth label estimates over the graphs \cite{smola2003kernels}, and  $\regfun(\cdot)$ is an $\mathcal{L}_2$ norm over the AGCN parameters that is used to avoid overfitting~\cite{goodfellow2016deep}. Finally, the $\mathcal{L}_1$ norm in the third regularizer encourages learning sparse mixing coefficients, and hence it promotes activating only a subset of edge-dithered graphs per $ l$. ED graphs with a large number of perturbed edges will result to a higher cost function in \eqref{eq:trainobj}. Hence, the learning algorithm will assign larger combining weights to non-perturbed topologies.  The backpropagation algorithm is employed to minimize \eqref{eq:trainobj}. 

To recap, aiming at a robust GCN architecture, we introduce a novel edge-dithering module that generates probabilistically  auxiliary graphs. These graphs are processed by a robust AGCN architecture that: combines features within neighborhoods originating from the different graphs; adapts to each graph by aggregating the learned features with $\underline{\mathbf{R}}$; uses a simple but versatile residual tensor  mapping \eqref{eq:residuallayer}; and employs smoothness and sparsity promoting graph-based regularizers; see also Fig.~\ref{fig:grnn}. 
		
\section{Experiments}
The AGCN is tested  with $\nbrlayers=3$, $ P^{(1)}=64$,	$ P^{(2)}=8$, and  $ P^{(w)}=K$. The regularization parameters $\{\regpargraphsmooth,\regparsmooth,\regparsparse\}$ are chosen based on the performance of the AGCN in the validation set for each experiment. For the training stage, an ADAM optimizer with learning rate 0.005 was employed \cite{kingma2015adam}, for 300 epochs with early stopping at 60 epochs. The goal here is to provide tangible answers to the following research questions.
\begin{itemize}
		\item[\textbf{RQ1}.] How does AGCN compare to state-of-the-art methods for SSL over multi-relational graphs?
		\item[\textbf{RQ2}.] How robust is AGCN compared to GCN under noisy features, noisy edge weights, and random as well as adversarial edge perturbations?
		\item[\textbf{RQ3}.] How sensitive is AGCN to the parameters of the ED module $q_1,q_2$ and $I$)?
\end{itemize}
	
\subsection{Predicting multi-relational protein functions}
This section assesses the performance of the proposed AGCN when predicting protein functions over multiple graphs. For this experiment, the given network is multi-relational and no perturbations were considered, hence the ED module is not used. Protein-to-protein interaction networks relate two proteins via multiple cell-dependent relations and  protein classification seeks the unknown function of some proteins based on the functionality of a (small) subset of them~\cite{zitnik2017predicting}. Given a target function $y_n$ that is known on a subset of proteins $n\in\mathcal{L}$, known functions on all proteins summarized in $\mathbf{X}$, and the multi-relational protein networks $\underline{\mathbf{A}}$, the goal is to predict whether proteins in the unlabeled set $n\in\{\mathcal{V}\setminus \mathcal{L}\}$ are associated with the target function or not. Hence, the number of target classes in this application is simply $K=2$. In this setting, $\mathbf{A}_i$ represents the protein connectivity in the  $i$-th cell type. Examples of such cells include cerebellum, midbrain, or frontal lobe. Table \ref{tab:biodata} summarizes the dimensions of the three datasets used in our experiments.

\begin{table}[t]
		\hspace{0cm}
		\centering
		\caption{Protein-to-protein interaction datasets}
		\rowcolors[]{1}{white}{gray}
		\begin{tabular}{c c c c c}
			\hline
			\textbf {Dataset} &  {Nodes} $N$  &  {Features} $F$ & {Relations} $I$\\
			\hline\hline
			Generic cells  & 4,487 & 502 & 144\\
			Brain cells & 2,702 & 81 & 9  \\
			Circulation cells & 3,385 & 62 & 4
		\end{tabular}
		\vspace{0.0cm}
		\label{tab:biodata}
\end{table}
Next, AGCN is compared with the GCN~\cite{kipf2016semi} that is the single-relational alternative, and the Mune~\cite{ye2018robust} that is a state-of-the-art diffusion-based approach for SSL over multi-relational graphs. Since GCN only accounts for a single graph, GCN employs the graph $i$ that achieves the best results in the validation set. Furthermore, Mune does not account for feature vectors in the nodes of the graph. Hence, to facilitate fair comparison, we also employ our AGCN without using the feature vectors, i.e. $\mathbf{X}=\mathbf{I}_N$. Finally, since the classes are heavily unbalanced, we evaluate the performance of the various approaches using the macro F1 score for predicting the protein functions.\footnote{Accurate classifiers achieve macro F1 values close to 1.}
	
\begin{table}[h!]
		\centering
		\caption{Macro F1  for the brain cells dataset.}
		
		\rowcolors[]{1}{white}{gray}
		\begin{tabular}{c c c c c}
			\hline
			$|\mathcal{L}|$& 440 & 220 & 110& 55\\
			\hline
			\hline
			AGCN & \textbf{0.86} &\textbf{0.79}
			&\textbf{0.71}
			& \textbf{0.69}
			\\
			GCN & 0.49 &0.48 &
			0.48&
			0.47
			\\
			AGCN (No feat.) & 0.41&
			0.43
			& 0.41
			& 0.35
			\\
			Mune (No feat.) & 0.27&
			0.27
			&0.32
			&0.14
			\\\hline
		\end{tabular} \vspace{0.01cm}  \label{tab:brain}
\end{table}
\begin{table}[h!]
		\centering
		\caption{Macro F1 for the circulation cells dataset.}
		
		\rowcolors[]{1}{white}{gray}
		\begin{tabular}{c c c c c}
			\hline
			$|\mathcal{L}|$& 440 & 220 & 110& 55\\
			\hline
			\hline
			AGCN & \textbf{0.77} &\textbf{0.76}
			&\textbf{0.70}
			& \textbf{0.69}
			\\
			GCN & 0.48 &0.48 &
			0.48&
			0.47
			\\
			AGCN (No feat.) & 0.41&
			0.42
			& 0.40
			& 0.35
			\\
			Mune (No feat.) & 0.28&
			0.27
			&0.26
			&0.13
			\\\hline
		\end{tabular}
		\vspace{0.01cm}
		\label{tab:circu}
\end{table}
\begin{table}[h!]
		\centering
		\caption{Macro F1  for the generic cells dataset.} 
		
		\rowcolors[]{1}{white}{gray}
		\begin{tabular}{c c c c c}
			\hline
			$|\mathcal{L}|$& 440 & 220 & 110& 55\\
			\hline
			\hline
			AGCN & \textbf{0.70} &\textbf{0.66}
			&\textbf{0.60}
			& \textbf{0.58}
			\\
			GCN & 0.49 &0.48 &
			0.48&
			0.47
			\\
			AGCN (No feat.) & 0.40&
			0.44
			& 0.41
			& 0.43
			\\
			Mune (No feat.) & 0.28&
			0.25
			&0.24
			&0.13
			\\\hline
		\end{tabular}
		\vspace{0.01cm}
		\label{tab:gen}
\end{table}

Tables \ref{tab:brain}-\ref{tab:gen} report macro F1 values for the aforementioned approaches for varying number of observed (labeled) nodes $|\mathcal{L}|$. The results for all datasets demonstrate that: i) the macro F1 score improves as $|\mathcal{L}|$ increases; ii) AGCN, that judiciously combines the multiple-relations, outperforms the GCN by a large margin; and iii) for the case where nodal features are not used (bottom two rows of each table), AGCN outperforms the state-of-the-art Mune.
\begin{figure}[t]
		{\input{figs/ionosphereaccvsfeatsnr.tex}}
		~\hspace{-3cm}
		{\input{figs/ionosphereaccvsadjsnr.tex}}
		\caption{SSL classification accuracy of AGCN %on ionosphere
			with $|\mathcal{L}|=50$ for noisy features (left) and noisy graphs (right). 
		}
		\label{fig:robust}
\end{figure}
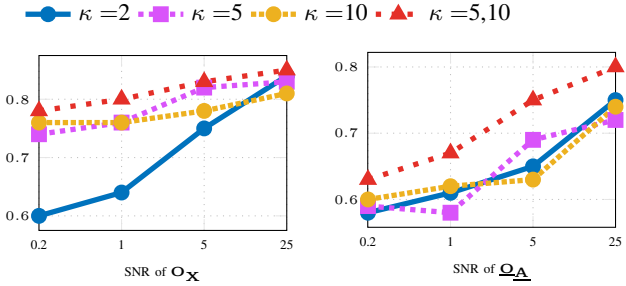
\subsection{Robustness to additive Gaussian input noise}
This section reports the performance of the proposed ED-AGCN architecture under noisy graphs or features. For this experiment,  the ionosphere dataset is considered, which contains $N=351$ data points with $F=34$ features that belong to $K=2$ classes \cite{Dua:2017}. In this case,  the graphs $\mathcal{G}_{i}$ are formed using $\kappa$-nearest neighbors graphs for different values of $\kappa$ (i.e., different number of neighbors). This method computes the link between $n$ and $n'$ based on the Euclidean distance of their features $\|\mathbf{x}_n-\mathbf{x}_n'\|_2^2$.
	
Oftentimes, the available topology and feature vectors might be corrupted with noise. Capturing this noise, $\underline{\mathbf{A}}$ and $\mathbf{X}$  are obtained as $\underline{\mathbf{A}}=\underline{\mathbf{A}}_{tr}+\shiftperturbation, \mathbf{X}=\mathbf{X}_{tr}+\featperturbation,$ where $\underline{\mathbf{A}}_{tr}$ and $\mathbf{X}_{tr}$ represent the \textit{true} multi-relational topology and features and $\shiftperturbation$ and $\featperturbation$ denote the corresponding additive perturbations. We draw  $\shiftperturbation$ and $\featperturbation$  from a zero mean white Gaussian distribution with specified signal to noise ratio (SNR). Since random additive noise is considered here, the ED module is not employed.
	
Fig. \ref{fig:robust} reports the SSL classification performance for the ionosphere dataset of AGCNs. The AGCN is tested with different values of $\kappa$, where $\kappa=5,10$ corresponds to a two relational graph, i.e. $I=2$. We deduce that multiple $\kappa$-nearest neighbors graphs lead to learning more robust representations of the data, which testifies to the merits of proposed multi-relational architecture. 
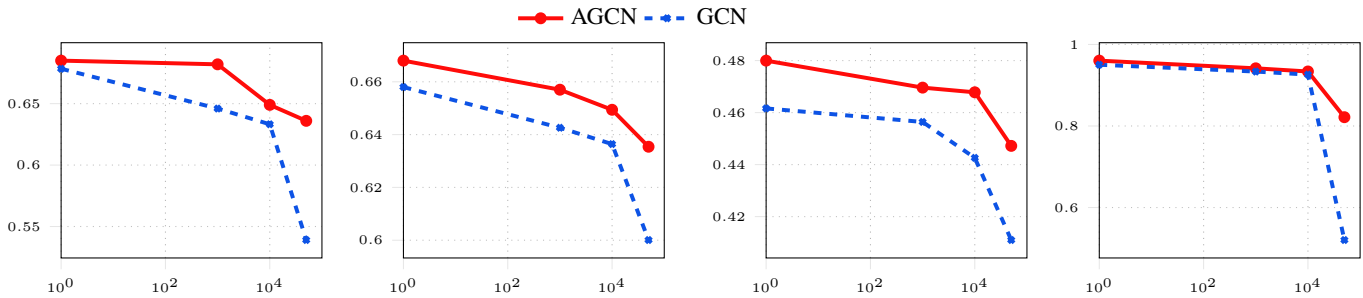
\begin{figure*}
		%\centering
		\hspace{-1cm}\input{figs/adlinkscora.tex}~\input{figs/adlinkspubmed.tex}\hspace{-1cm}~\input{figs/adlinkciteseer.tex}~\input{figs/adlinkspolblog.tex}
		\caption{Classification accuracy for increasing number of perturbed edges. (\textbf{Left}) Cora, (\textbf{Middle left}) Pubmed, (\textbf{Middle right}) Citeseer, (\textbf{Right} Polblogs).}
		\label{fig:adrandpert}
\end{figure*}
{\setlength\extrarowheight{2pt}
\begin{table}
			\caption{Classification accuracy  in percent for nodes in $\mathcal{T}$ for different numbers of attacked nodes.}
			
			\label{tab:results}
			%\resizebox{0.93\textwidth}{!}{ \renewcommand{\arraystretch}{0.6}
			\centering
			\begin{tabular}{@{}p{1cm}p{1cm}cccccc@{}}\cmidrule[\heavyrulewidth]{1-7}
				\multirow{3}{*}{\vspace*{8pt}\textbf{Dataset}}& \multirow{3}{*}{\vspace*{8pt}\textbf{Method}}&\multicolumn{4}{c}{Number of attacked nodes $|\mathcal{T}|$}\\\cmidrule{3-7}
				& & {\textsc{20}} & {\textsc{30}} & {\textsc{40}} & {\textsc{50}} & {\textsc{60}} 
				\\ \cmidrule{1-7}
				\multirow{2}{*}{\rotatebox{0}{\hspace*{-0pt}\emph{Citeseer}} }
				& \textsc{GCN}  & 60.49 & 56.00 &  61.49 & 56.39 & \textbf{58.99}  \\ 
				& \textsc{AGCN} & \textbf{70.99}& \textbf{56.00} & \textbf{61.49} & \textbf{61.20}  & 58.66 \\     
				\cmidrule{1-7}
				\multirow{2}{*}{\rotatebox{0}{\hspace*{-4pt}\emph{Cora}}} 
				& \textsc{GCN}  & 76.00 & 74.66 &  76.00 & 62.39 & 73.66  \\ 
				& \textsc{AGCN} & \textbf{78.00} & \textbf{82.00} & \textbf{84.00} & \textbf{73.59}  & \textbf{74.99}\\     
				\cmidrule{1-7}
				\multirow{2}{*}{\rotatebox{0}{\hspace*{-4pt}\emph{Pubmed}}} 
				& \textsc{GCN}  & \textbf{74.00} & 71.33 &  68.99 & 66.40 & 69.66  \\ 
				& \textsc{AGCN} & 72.00 & \textbf{75.36} & \textbf{71.44} & \textbf{68.50}  & \textbf{74.43}    \\     
				\cmidrule{1-7}
				\multirow{2}{*}{\rotatebox{0}{\hspace*{-4pt}\emph{Polblogs}}} 
				& \textsc{GCN}  & \textbf{85.03} & 86.00 &  84.99 & 78.79 & 86.91  \\ 
				& \textsc{AGCN} & 84.00 & \textbf{88.00} & \textbf{91.99} & \textbf{78.79}  & \textbf{92.00}   \\   
				\cmidrule[\heavyrulewidth]{1-7}
			\end{tabular}%}
\end{table}
}\setlength\extrarowheight{0pt}
\subsection{Robustness to Bernoulli noise on edges}
This experiment tests our architecture with four network datasets~\cite{sen2008collective}: ``Cora'' ($N=2708,K=7,|\mathcal{L}|=140$), ``Citeseer'' ($N=3327,K=6,|\mathcal{L}|=120$) and ``Pubmed'' ($N=19717,K=3,|\mathcal{L}|=30$) are citation graphs, while ``Polblogs'' ($N=1224,K=2,|\mathcal{L}|=24$) is a political blog network. To facilitate comparison, we reproduce the same experimental setup than in  \cite{kipf2016semi}, i.e., the same split of the data in train, validation, and test sets. To study the effect of graph perturbations on the neural network architecture, the feature vectors of the citation datasets are not used. Notice that our robust GCN architecture can be applied even in the absence of nodal features, whereas existing approaches are not directly applicable \cite{wu19adv}. For this experiment, the perturbed graph $\bar{\mathbf{A}}$ is generated by inserting new edges in the original graphs between a random pair of nodes $n,n'$ that are not connected in $\mathbf{A}$, i.e. $A_{n,n'}=0$. The added edges can be regarded as drawn from Bernoulli distribution. AGCN utilizes the multiple graphs generated via the ED module with $I=10$ samples, $q_1=0.9$, and $q_2=1$ since no edge is deleted in $\bar{\mathbf{A}}$.
	
Fig.~\ref{fig:adrandpert} demonstrates the classification accuracy of the  GCN~\cite{kipf2016semi} compared to the proposed AGCN as the number of perturbed edges is increasing. Evidently, AGCN utilizes the novel ED module, and achieves robust SSL compared to GCN. Surprisingly, even when no edges are perturbed, AGCN outperforms GCN. This observation may be attributed to noisy links in the original graphs, which hinder classification perfomance. Furthermore, SSL performance of GCN significantly degrades as the number of perturbed edges increases, which suggests that GCN is challenged even by ``random attacks''.

\subsection{Robustness to adversarial attacks on edges}
The original graphs in Cora, Citeseer, Pubmed, and Polblogs were perturbed using the adversarial setup in~\cite{zugner18adv}, where structural attacks are effected on attributed graphs. These attacks perturb connections adjacent to  $\mathcal{T}$ a set of targeted nodes by adding or deleting edges~\cite{zugner18adv}. Our ED module uses $I=10$ sampled graphs with $q_1=0.9$, and $q_2=0.999$.  For this experiment, 30\% of the nodes are used for training, 30\% for validation and 40\% for testing.\footnote{The nodes in $\mathcal{T}$ are in the testing set.}
	
Table \ref{tab:results} reports the classification accuracy of GCN and AGCN for different number of attacked nodes ($|\mathcal{T}|$). Different from Fig.~\ref{fig:adrandpert} where the classification accuracy over the test set is reported, Table \ref{tab:results}  reports the classification accuracy over the set of attacked nodes $\mathcal{T}$. It is observed that the proposed AGCN is more robust relative to GCN under adversarial attacks~\cite{zugner18adv}. This finding justifies the use of the novel ED in conjunction with the AGCN that judiciously selects extracted features originating from non-corrupted neighborhoods.

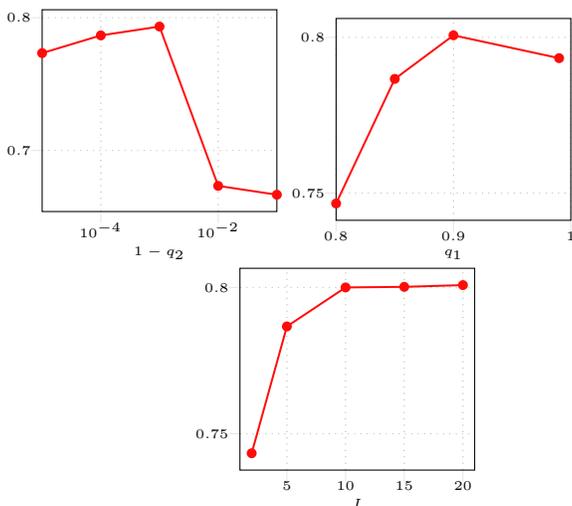
\begin{figure}[t]
		{\input{figs/agcnq2per.tex}}~\hspace{-0.1cm}{\input{figs/agcnp2per.tex}}
		
		\centering\hspace{0.0cm}\vspace{0.1cm}{\input{figs/agcnI2per.tex}}\vspace{-0.0cm}
		\caption{SSL classification accuracy of AGCN under varying edge creation prob. $q_1$, edge deletion prob. $q_2$, and number of samples $I$.
		}
		\label{fig:robustsens}
\end{figure}

\noindent\textbf{Parameter sensitivity analysis}.
Fig. \ref{fig:robustsens} includes sensitivity of the AGCN to varying parameters of the ED module for the experiment in Table \ref{tab:results} with the Cora and $|\mathcal{T}|=30$.  It is observed that the AGCN's performance is relative smooth for certain ranges of the parameters. In accordance with Remark 2, notice that even for small $I$  AGCN's performance is increased significantly. 
	
\section{Conclusions}
This work advocates a novel deep learning approach to robust SSL over perturbed graphs. It relies on random dithering applied to edges with probabilities selected to restore a node's original neighborhood with high probability. The auxiliary edge-dithered graphs are combined and jointly exploited by an adaptive GCN. The latter assigns larger combining weights to learned features extracted from graph neighborhoods without perturbed edges. Experiments demonstrate the performance gains of AGCN in the presence of noisy features, noisy edge weights, and random as well as adversarial edge perturbations.

\small    
\bibliographystyle{IEEEtran}
\bibliography{my_bibliography}
\noindent
	\appendix
	\onecolumn
	{\Large{\centering\textbf{Edge Dithering for Robust Adaptive Graph Convolutional Networks\\~~~~~~~~~~~~~~~~~~~~~~~~~~~~~~~~~~~~~~~~~~~~~~~~~~~~~~~~~~~~~Supplementary material}}}
	\section{Proof of Remark 2}
	
	\begin{myremark}
		With high probability there exists $\mathcal{G}_i$ such that a perturbed edge will be restored to its initial value. This means that there exist a graph $i$ such that $A_{n,n',i}=A_{n,n'}$.
		Since, each $\mathcal{G}_i$ is independently drawn, it holds that \footnote{Equations ($k$) with $k\le13$ correspond to the orginal manuscript while ($k$) with $k>13$ correspond to the supplementary material).}
		\begin{align}
		{\rm Pr}\bigg(\wbigcup_{i=1}^I(A_{n,n',i}=0)\Big|\bar{A}_{n,n'}=1,A_{n,n'}=0\bigg)=1-q_1^I\\
		{\rm Pr}\bigg(\wbigcup_{i=1}^I(A_{n,n',i}=1)\Big|\bar{A}_{n,n'}=0,A_{n,n'}=1\bigg)=1-q_2^I
		\end{align}
	\end{myremark}
	
	\begin{myremark}
		With high probability there exists $\mathcal{G}_i$ which will recover the original neighborhood structure of a node, i.e. $\mathcal{N}_n^{(i)}=\mathcal{N}_n$. 
	\end{myremark}
	
	\textbf{Proof}.
	The two neighborhood structures will be the same $\mathcal{N}_n^{(i)}=\mathcal{N}_n$ if and only if $A_{n,n',i}=A_{n,n'}, \forall n'$. For any edge $A_{n,n',i}$ there are 4 scenarios
	\begin{itemize}
		\item[1)] $\check{A}_{n,n'}=1$ and $A_{n,n'}=1$
		\item[2)] $\check{A}_{n,n'}=1$ and $A_{n,n'}=0$
		\item[3)] $\check{A}_{n,n'}=0$ and $A_{n,n'}=1$
		\item[4)] $\check{A}_{n,n'}=0$ and $A_{n,n'}=0$
	\end{itemize}
	It further holds that
	\begin{align}
	{\rm Pr} \bigg(\wbigcup_{i=1}^I(A_{n,n',i}=0)\Big|\bar{A}_{n,n'}=1,A_{n,n'}&=0 \bigg)=1-q_1^I\\
	{\rm Pr} \bigg(\wbigcup_{i=1}^I(A_{n,n',i}=1)\Big|\bar{A}_{n,n'}=0,A_{n,n'}&=1 \bigg)=1-q_2^I\\
	{\rm Pr} \bigg(\wbigcup_{i=1}^I(A_{n,n',i}=0)\Big|\bar{A}_{n,n'}=1,A_{n,n'}&=1 \bigg)=1-(1-q_1)^I\\
	{\rm Pr} \bigg(\wbigcup_{i=1}^I(A_{n,n',i}=1)\Big|\bar{A}_{n,n'}=0,A_{n,n'}&=0 \bigg)=1-(1-q_2)^I
	\end{align}
	Without loss of generality assume for the $N$ connections $\{A_{n,n'}\}_{n'=1}^N$ the events 1)-4) appear with the following frequency $\kappa,\lambda,\mu,\nu$. Since sampling each edge of the graph is done independently across edges and across draws we arrive to the following
	\begin{align}
	{\rm Pr}\bigg(\wbigcup_{i=1}^I(\mathcal{N}_n^{(i)}=\mathcal{N}_n)\bigg)=\left(1-(1-q_1)^I\right)^\kappa\left(1-q_1^I\right)^\lambda\left(1-q_2^I\right)^\mu\left
	(1-(1-q_2)^I\right)^\nu
	\end{align}
	%\noindent
\end{document}

%% file: include.tex
\usepackage{graphicx}

\usepackage{subcaption}              %subfigures with \mbox and \subfigure[]

\usepackage[utf8]{inputenc}

\usepackage{amsfonts}
\usepackage{amsmath}
\usepackage{mathtools}
\usepackage{amssymb}

\usepackage{bm}

\usepackage{color,verbatim}
\usepackage{multirow}
\usepackage{accents}

\usepackage{theoremref}
\usepackage{flushend}

\usepackage{url}

\newcounter{rulecounter}
\newcommand{\resetrule}{ \setcounter{rulecounter}{0}}
\resetrule

\newsavebox{\selvestebox}
\newenvironment{colbox}[1]
  {\newcommand\colboxcolor{#1}%
   \begin{lrbox}{\selvestebox}%
   \begin{minipage}{\dimexpr\columnwidth-2\fboxsep\relax}}
  {\end{minipage}\end{lrbox}%
   \begin{center}
   \colorbox{\colboxcolor}{\usebox{\selvestebox}}
   \end{center}}

\definecolor{orange}{rgb}{1,0.8,0}
\definecolor{gray}{rgb}{.9,0.9,0.9}
\definecolor{darkgray}{rgb}{.3,0.3,0.3}
\definecolor{darkblue}{rgb}{.1,0.0,0.3}
\definecolor{lightblue}{rgb}{0.7,0.7,1}
\definecolor{lightred}{rgb}{1,0.7,.7}
\definecolor{purple}{RGB}{204,153,255}
\definecolor{lightgray}{rgb}{.95,0.95,0.95}
\definecolor{lightgreen}{rgb}{0.3,0.5,0.3}
\definecolor{darkgreen}{rgb}{0.05,0.3,0.05}

 \newcommand{\define}{:=}
\newtheorem{myproposition}{Proposition}
\newtheorem{myremark}{Remark}
\newtheorem{myproblemstatement}{Problem Statement}
\newtheorem{mylemma}{Lemma}
\newtheorem{mytheorem}{Theorem}
\newtheorem{mydefinition}{Definition}
\newtheorem{mycorollary}{Corollary}

\usepackage{pgfplots}
\pgfplotsset{compat=newest}
\usetikzlibrary{plotmarks}
\usetikzlibrary{positioning}
\usetikzlibrary{arrows.meta}
\usepgfplotslibrary{patchplots}
\usepackage{grffile}

\pgfplotsset{plot coordinates/math parser=false}
\newlength\mywidth
\newlength\myheight
\newlength\mywidths
\newlength\myheights
\newlength\mywidthss
\newlength\myheightss
\definecolor{mycolor1}{rgb}{0.00000,0.44700,0.74100}%
\definecolor{mycolor2}{rgb}{0.85000,0.32500,0.9800}%
\definecolor{mycolor3}{rgb}{0.92900,0.69400,0.12500}%
\definecolor{mycolor4}{rgb}{0.89400,0.18400,0.15600}%
\definecolor{mycolor5}{rgb}{0.46600,0.67400,0.18800}%
\definecolor{mycolor6}{rgb}{0.30100,0.74500,0.93300}%
\definecolor{mycolor7}{rgb}{0.63500,0.07800,0.18400}%

\definecolor{AGNN}{rgb}{1,0.04700,0.04100}%
\definecolor{GCN}{rgb}{0.05000,0.32500,0.89800}%
\definecolor{AGNNnf}{rgb}{1,0.2,0.119800}%
\definecolor{Mune}{rgb}{0.2,0.2700,1}%

\definecolor{TightHann}{rgb}{0.92900,0.69400,0.12500}%
\definecolor{pDiffScater}{rgb}{0.49400,0.18400,0.55600}%
\definecolor{pMonicCubic}{rgb}{0.46600,0.67400,0.18800}%
\definecolor{pTightHann}{rgb}{0.30100,0.74500,0.93300}%

\pgfplotscreateplotcyclelist{colorlist}{%
color=mycolor1,line width=2.0pt,
solid, every mark/.append style={solid, fill=mycolor1}, mark=*\\%
color=mycolor2,line width=2.0pt,dotted, every mark/.append style={solid, fill=mycolor2}, mark=square*\\%
color=mycolor3,line width=2.0pt,densely dotted, every mark/.append style={solid, fill=mycolor3}, mark=otimes*\\%
color=mycolor4,line width=2.0pt,loosely dotted, every mark/.append style={solid, fill=mycolor4}, mark=triangle*\\%
color=mycolor5,line width=2.0pt,
dashed, every mark/.append style={solid, fill=mycolor5},mark=diamond*\\%
color=mycolor6,line width=2.0pt,loosely dashed, every mark/.append style={solid, fill=mycolor6},mark=*\\%
color=mycolor7,densely dashed, every mark/.append style={solid, fill=mycolor7},line width=2.0pt,mark=square*\\%
dashdotted, every mark/.append style={solid, fill=mycolor7},mark=otimes*\\%
dashdotdotted, every mark/.append style={solid},mark=star\\%
densely dashdotted,every mark/.append style={solid, fill=gray},mark=diamond*\\%
}
\pgfplotscreateplotcyclelist{my black white}{%
solid, every mark/.append style={solid, fill=gray}, mark=*\\%
dotted, every mark/.append style={solid, fill=gray}, mark=square*\\%
densely dotted, every mark/.append style={solid, fill=gray}, mark=otimes*\\%
loosely dotted, every mark/.append style={solid, fill=gray}, mark=triangle*\\%
dashed, every mark/.append style={solid, fill=gray},mark=diamond*\\%
loosely dashed, every mark/.append style={solid, fill=gray},mark=*\\%
densely dashed, every mark/.append style={solid, fill=gray},mark=square*\\%
dashdotted, every mark/.append style={solid, fill=gray},mark=otimes*\\%
dashdotdotted, every mark/.append style={solid},mark=star\\%
densely dashdotted,every mark/.append style={solid, fill=gray},mark=diamond*\\%
}
\usetikzlibrary{decorations.pathreplacing}
\newcommand{\xlabelfontsize}{\small}

\newcommand{\legendfontsize}{\normalsize}
\newcommand{\ticklabelfontsize}{\normalsize}
\definecolor{lavander}{cmyk}{0,0.48,0,0}
\definecolor{violet}{cmyk}{0.79,0.88,0,0}
\definecolor{burntorange}{cmyk}{0,0.52,1,0}
\definecolor{mygreen}{rgb}{0,1,0}%
\definecolor{myred}{rgb}{1,0,0}%

\newcommand{\mylinewidth}{1.5}
\newcommand{\markwidth}{1.5}
\newcommand{\distbetweenlayers}{1.8}
\tikzstyle{nnblock}=[draw,rectangle,  rounded corners,text=black,
minimum width=30pt,minimum height=60pt]
\tikzstyle{esl}=[draw,rectangle,  rounded corners,text=black,
minimum width=60pt,minimum height=30pt]
\tikzstyle{grnn}=[draw,rectangle,  rounded corners,text=black,
minimum width=30pt,minimum height=15pt]
\tikzstyle{graphsampler}=[draw,rectangle,  fill=orange,fill opacity=0.1, rounded corners,text=orange,
minimum width=30pt,minimum height=15pt]
\tikzstyle{outblock}=[draw,rectangle, color=red, rounded corners,text=black,
minimum width=10pt,minimum height=60pt]
\tikzstyle{nam}=[rectangle, fill=orange,fill opacity=0.1, rounded corners,text=orange,
minimum width=8pt,minimum height=58pt]
\tikzstyle{gam}=[fill=blue,fill opacity=0.1, rectangle,  rounded corners,text=black,
minimum width=8pt,minimum height=58pt]
\tikzstyle{fam}=[fill=green,fill opacity=0.1, rectangle,  rounded corners,text=black,
minimum width=8pt,minimum height=58pt]

\tikzstyle{peers}=[draw,circle,bottom color=myred,
                  top color= white, text=violet,minimum width=8pt]
\tikzstyle{superpeers}=[draw,circle, left color=mygreen,
                       text=violet,minimum width=8pt]

\tikzstyle{peers3}=[draw,circle,bottom color=myred,%lavander,
                  top color= white, text=violet,minimum width=12pt]
\tikzstyle{superpeers3}=[draw,circle, left color=mygreen,%orange,
                       text=violet,minimum width=12pt]
\tikzstyle{peers2}=[draw,circle,bottom color=myred,
                  top color= white, text=violet,inner sep=0pt,minimum size=0.1cm]
\tikzstyle{superpeers2}=[draw,circle, left color=mygreen,
                       text=violet,inner sep=0pt,minimum size=0.1cm]
                    
\tikzstyle{peers1}=[draw,circle,bottom color=myred,%lavander,
                  top color= white, text=violet,minimum width=4pt]
\tikzstyle{superpeers1}=[draw,circle, left color=mygreen,%orange,
                       text=violet,minimum width=4pt]
\tikzstyle{legendsp}=[rectangle, draw,  rounded corners,
                     thin,bottom color=mygreen, top color=white,
                     text=black, minimum width=2cm]
\tikzstyle{legendp}=[rectangle, draw,  rounded corners, thin,
                     bottom color=myred, top color= white,
                     text= black, minimum width= 2cm]
\tikzstyle{legend_general}=[rectangle, rounded corners, thin,
                           burntorange, fill= white, draw, text=violet,
                           minimum width=2.5cm, minimum height=0.8cm]
\usetikzlibrary{shapes.misc, positioning}

\usetikzlibrary{calc,trees}

\newcommand{\convexpath}[2]{
[   
    create hullnodes/.code={
        \global\edef\namelist{#1}
        \foreach [count=\counter] \nodename in \namelist {
            \global\edef\numberofnodes{\counter}
            \node at (\nodename) [draw=none,name=hullnode\counter] {};
        }
        \node at (hullnode\numberofnodes) [name=hullnode0,draw=none] {};
        \pgfmathtruncatemacro\lastnumber{\numberofnodes+1}
        \node at (hullnode1) [name=hullnode\lastnumber,draw=none] {};
    },
    create hullnodes
]
($(hullnode1)!#2!-90:(hullnode0)$)
\foreach [
    evaluate=\currentnode as \previousnode using \currentnode-1,
    evaluate=\currentnode as \nextnode using \currentnode+1
    ] \currentnode in {1,...,\numberofnodes} {
-- ($(hullnode\currentnode)!#2!-90:(hullnode\previousnode)$)
  let \p1 = ($(hullnode\currentnode)!#2!-90:(hullnode\previousnode) - (hullnode\currentnode)$),
    \n1 = {atan2(\y1,\x1)},
    \p2 = ($(hullnode\currentnode)!#2!90:(hullnode\nextnode) - (hullnode\currentnode)$),
    \n2 = {atan2(\y2,\x2)},
    \n{delta} = {-Mod(\n1-\n2,360)}
  in 
    {arc [start angle=\n1, delta angle=\n{delta},radius=#2]}
}
-- cycle
}

%% file: my_definitions.tex
\newcommand{\featvec}{\mathbf{x}}
\newcommand{\dataentry}{X}
\newcommand{\datamatrix}{\mathbf{\dataentry}}
\newcommand{\datatensor}{\underline{\datamatrix}}

\newcommand{\nbrnodes}{N}
\newcommand{\shiftentry}{A}

\newcommand{\shiftmat}{\mathbf{\shiftentry}}
\newcommand{\shiftind}{i}
\newcommand{\nbrshifts}{I}

\newcommand{\nbrlayers}{L}

\newcommand{\graph}{\mathcal{G}}

\newcommand{\regparsmooth}{\mu_2}
\newcommand{\regparsparse}{\lambda}
\newcommand{\regpargraphsmooth}{\mu_1}

\newcommand{\nodeshiftnot}[2]{_{#1#2}}

\newcommand{\layernot}[1]{^{(#1)}}
\newcommand{\layerind}{l}
\newcommand{\layfunc}{f}

\newcommand{\nodeind}{n}

\newcommand{\shifttensor}{\underline{\shiftmat}}

\newcommand{\outputlaymat}{\check{\mathbf{Z}}}

\newcommand{\outputlaytensor}{\underline{\outputlaymat}}

\newcommand{\diffusedfeatvec}{\mathbf{h}}
\newcommand{\graphadaptivec}{\mathbf{g}}

\newcommand{\nbrclasses}{K}

\newcommand{\predictionmat}{\hat{\mathbf{Y}}}

\newcommand{\shiftperturbation}{\underline{\mathbf{O}}_{\shifttensor}}
\newcommand{\featperturbation}{\mathbf{O}_{\datamatrix}}

\newcommand{\regfun}{\rho}

%% file: figs/sampledGraph.tex
\begin{tikzpicture}[%
        scale=0.8, every node/.style={scale=0.8},
        block/.style={rectangle, fill=gray, text centered, minimum height=8mm, rounded corners=1mm}
        ,my line style1/.style={ draw=red,thick}
        ,my line style2/.style={ draw=blue,thick}
        ,my line style3/.style={ draw=black,thick}
        ,every label/.append style = {label distance=1pt, inner sep=1pt, 
                             align=left, font=\tiny}
        %,every edge/.style={my line style}
        ,my edge/.style={my line style, mylabel=#1}
        ,my curved edge/.style n args={4}{my line style, out=#1, in=#2, looseness=#3, mylabel=#4}
        ,edge1/.style={my line style1,dashed}
        ,mylabel1/.style={my label style{#1}}
        ,edge2/.style={my line style2, densely dotted}
        ,edge3/.style={my line style3}
        ,mylabel/.style={edge node={node[my label style]{#1}}}]
  % Place super peers and connect them
\newcommand{\disttopgraphx}{-2}
\newcommand{\disttopgraphy}{+2}
\foreach \place/\name in {{(5+\disttopgraphx,-0.5+\disttopgraphy)/ar0}, {(5.5
+\disttopgraphx,0.3+\disttopgraphy)/br0}, {(5+\disttopgraphx,0.7+\disttopgraphy)/cr0}, {(4.5+\disttopgraphx,0.7+\disttopgraphy)/dr0}}
    \node[superpeers1] (\name) at \place {};
\foreach\place/\name  in {{(6+\disttopgraphx,-0.5+\disttopgraphy)/ad0}, {(6.5+\disttopgraphx,0+\disttopgraphy)/bd0}, {(6+\disttopgraphx,0.7+\disttopgraphy)/cd0}, {(7+\disttopgraphx,0.7+\disttopgraphy)/dd0}}
\node[peers1] (\name) at \place {};

\foreach \distxbetweengraphs\distybetweengraphs\graphid in {{0/-3/1},{3.5/-3/2},{7/-3/3}}
  \foreach \place/\name in {{(0+\distxbetweengraphs,-0.5+\distybetweengraphs)/ar\graphid}, {(0.5+\distxbetweengraphs,0.3+\distybetweengraphs)/br\graphid}, {(0+\distxbetweengraphs,0.7+\distybetweengraphs)/cr\graphid}, {(-0.5+\distxbetweengraphs,0.7+\distybetweengraphs)/dr\graphid}}
    \node[superpeers1] (\name) at \place {};
  
   %
   % Place normal peers
  \foreach \distxbetweengraphs\distybetweengraphs\graphid in  {{0/-3/1},{3.5/-3/2},{7/-3/3}}
  \foreach\place/\name  in {{(1+\distxbetweengraphs,-0.5+\distybetweengraphs)/ad\graphid}, {(1.5+\distxbetweengraphs,0+\distybetweengraphs)/bd\graphid}, {(1+\distxbetweengraphs,0.7+\distybetweengraphs)/cd\graphid}, {(2+\distxbetweengraphs,0.7+\distybetweengraphs)/dd\graphid}}
    \node[peers1] (\name) at \place {};

   \foreach \source/\dest in {ar0/bd0, ar0/cr0, ad0/dd0,bd0/cd0,  br0/cr0, cr0/dr0,
   cd0/dd0,dr0/ad0}
    \path (\source) edge[edge3] (\dest);
    \path (ar0) edge[edge1] (ad0);
    \path (br0) edge[edge1] (bd0);

     \foreach \source/\dest in {ar1/bd1, ad1/dd1,bd1/cd1,  br1/cr1, cr1/dr1,
   cd1/dd1,dr1/ad1}
    \path (\source) edge[edge3] (\dest);
    \path (br1) edge[edge1] (bd1);

  \foreach \source/\dest in {ar2/bd2, ar2/cr2, cr2/dr2,ad2/dd2,br2/cr2,
   cd2/dd2,dr2/ad2}
    \path (\source) edge[edge3] (\dest);
    %\path (ar2) edge[edge1] (ad2);
    
    \foreach \source/\dest in {ar3/bd3, ar3/cr3, ad3/dd3, cr3/dr3,
   cd3/dd3,dr3/ad3}
    \path (\source) edge[edge3] (\dest);
    \path (ar3) edge[edge1] (ad3);
    \path (br3) edge[edge1] (bd3);
    \node[esl] at (3.5,0) (esl1) {\normalsize ED$(q_1,q_2)$};
  %labels 
  \node[label={[font=\normalsize]0: Corrupted graph $\bar{\mathcal{G}}$}] at (4.3+\disttopgraphx,1.5+\disttopgraphy) (cg) {};
   \node[] at (5.5+\disttopgraphx,-0.5+\disttopgraphy) (cg0) {};
  \node[] at (0.5,-2.3) (cg1) {};
  \node[] at (4.2,-2.3) (cg2) {};
  \node[]at (7.8,-2.3) (cg4) {};
  \node[label={[font=\normalsize]0: \large{$\cdots$}}] at (4.2,-1.5) (cg6) {};
  \node[label={[font=\normalsize]0: \Huge{$\cdots$}}] at (5.5,-3) (cg3) {};
  \draw[double,->] (cg0) --  (esl1);
  \draw[dashed,->] (esl1) -- node [above] {$\mathcal{G}_1$} (cg1);
   \draw[dashed,->] (esl1) -- node [left] {$\mathcal{G}_2$} (cg2);
    \draw[dashed,->] (esl1) -- node [above] {$\mathcal{G}_I$} (cg4);
  \node at (7.5+\disttopgraphx,1.7) (nA1) {};
    \node[label={[font=\normalsize]0: : True links}] at (8.5+\disttopgraphx,1.7) (nB1) {};
    \path (nA1) edge[edge3] (nB1);
  \node at (7.5+\disttopgraphx,1.3) (nA2) {};
    \node[label={[font=\normalsize]0: : Corrupted links}] at (8.5+\disttopgraphx,1.3) (nB2) {};
    \path (nA2) edge[edge1] (nB2);
     \node[superpeers,label={[font=\normalsize]0: : Democrat}] (d) at (8.4+\disttopgraphx,2.8) {};
    \node[peers,label={[font=\normalsize]0: : Republican}] (r) at (8.4+\disttopgraphx,2.3) {};
    
\end{tikzpicture}

%% file: figs/neighboragrmodule.tex
\begin{tikzpicture}[%
 block/.style={rectangle, fill=gray, text centered, minimum height=8mm, rounded corners=1mm}
        ,my line style1/.style={ draw=purple,dashed,thick}
        ,my line style2/.style={ draw=blue,densely dotted,thick}
        ,my line style3/.style={ draw=black,dashdotted,thick}
        ,my line style5/.style={ draw=black,thick}
        ,my line style4/.style={ draw=cyan,dotted, very thick}
        ,my line style6/.style={ draw=red,dashed, very thick}
        ,every label/.append style = {label distance=1pt, inner sep=1pt, 
                             align=left, font=\tiny}
        %,every edge/.style={my line style}
        ,my edge/.style={my line style, mylabel=#1}
        ,my curved edge/.style n args={4}{my line style, out=#1, in=#2, looseness=#3, mylabel=#4}
        ,edge1/.style={my line style5}
        ,mylabel1/.style={my label style{#1}}
        ,edge2/.style={my line style5}
        ,edge3/.style={my line style5}
         ,edge4/.style={my line style6}
        ,mylabel/.style={edge node={node[my label style]{#1}}}]
  % Place super peers and connect them
\foreach \distxbetweengraphs\distybetweengraphs\graphid in {{0/3/0},{4/3/1},{2/0/2}}
  \foreach \place/\name in {{(0+\distxbetweengraphs,-0.5+\distybetweengraphs)/ar\graphid}, {(0.5+\distxbetweengraphs,0.1+\distybetweengraphs)/br\graphid}, {(0+\distxbetweengraphs,1+\distybetweengraphs)/cr\graphid}, {(-0.5+\distxbetweengraphs,0.6+\distybetweengraphs)/dr\graphid},
           {(-0.5+\distxbetweengraphs,0+\distybetweengraphs)/er\graphid}}
    \node[superpeers3] (\name) at \place {};
  
   %
   % Place normal peers
  \foreach \distxbetweengraphs\distybetweengraphs\graphid in {{0/3/0},{4/3/1},{2/0/2}}
  \foreach\place/\name  in {{(1+\distxbetweengraphs,-0.5+\distybetweengraphs)/ad\graphid}, {(1.5+\distxbetweengraphs,0+\distybetweengraphs)/bd\graphid}, {(0.6+\distxbetweengraphs,1.2+\distybetweengraphs)/cd\graphid}, {(2+\distxbetweengraphs,1+\distybetweengraphs)/dd\graphid},
                        {(2+\distxbetweengraphs,0+\distybetweengraphs)/ed\graphid}}
    \node[peers3] (\name) at \place {};

   \foreach \source/\dest in { ar0/cr0, bd0/cd0,  br0/cr0, cr0/dr0,ad0/ed0,dr0/er0,
   cd0/dd0,dr0/ad0}
    \path (\source) edge[edge1] (\dest);
    \foreach \source/\dest in {ar1/er1,bd1/dd1, cr1/dr1,bd1/ed1,
   cr1/dr1,dr1/bd1,ed1/ad1}
    \path (\source) edge[edge2] (\dest);
    \foreach \source/\dest in {ar2/er2,bd2/dd2,bd2/ed2, ad2/ed2,dr2/er2,cr2/dr2,
   cd2/dd2,dr2/ad2,br2/ar2}
    \path (\source) edge[edge3] (\dest);
    \foreach \source/\dest in {ar0/bd0,ad0/dd0,dd1/er1,cd2/ed2, cd1/ed1,br1/bd1}
    \path (\source) edge[edge4] (\dest);
    %%%%
    \node[label={[font=\footnotesize]0:\textbf{\color{orange}Neighborhood aggregation module}}] at (0.6,5) (title) {};
  %\node at (-0.8,3.6) (nA2) {};
  %  \node[label={[font=\footnotesize]0: : $K=2$-hop neighborhood}] at (0,3.6) (nB2) {};
  %  \path (nA2) edge[edge4] (nB2);
    %%%%%%%%
    \node at (0,2.35) (s11) {};
    \node at (-0.5,1.6) (s21) {\footnotesize{\color{purple}$\diffusedfeatvec\nodeshiftnot{\nodeind}{1}\layernot{0}$}};
    \draw[->,my line style4,to path={-| (\tikztotarget)}]
        (s11) edge (s21);
     \node at (6,2.7) (s12) {};
    \node at (5.8,1.8) (s22) {\footnotesize{\color{blue}$\diffusedfeatvec\nodeshiftnot{\nodeind}{2}\layernot{0}$}};
    \draw[->,my line style4,to path={-| (\tikztotarget)}]
        (s12) edge (s22);
    \node at (1.5,-0.4) (s13) {};
    \node at (0.8,0.2) (s23) {\footnotesize{\color{black}$\diffusedfeatvec\nodeshiftnot{\nodeind}{3}\layernot{0}$}};
    \draw[->,my line style4,to path={-| (\tikztotarget)}]
        (s13) edge (s23);
   %%%%%%%
   %\draw[my line style4] \convexpath{er0,dr0,cr0,br0,ar0}{8pt};
   %\draw[my line style4] \convexpath{br1,dr1,dd1,ed1,bd1}{8pt};
   %\draw[my line style4] \convexpath{er2,br2,ar2}{8pt};
   \newcommand{\myfontnow}{\tiny}
   \node at (0.5,3.1) (nA) {\myfontnow $\featvec_\nodeind$};
   \node at (4.5,3.1) (nB) {\myfontnow $\featvec_\nodeind$};
   \node at (2.5,0.1) (nC) {\myfontnow $\featvec_\nodeind$};
      \node at (1,2.5) (nA) {\myfontnow $\featvec_{\nodeind_{\scalebox{.8} 9}}$};
   \node at (5,2.5) (nB) {\myfontnow $\featvec_{\nodeind_{\scalebox{.8}9}}$};
   \node at (3,-0.5) (nC) {\myfontnow $\featvec_{\nodeind_{\scalebox{.8}9}}$};
   \node at (0,2.5) (nA) {\myfontnow $\featvec_{\nodeind_{\scalebox{.8}1}}$};
   \node at (4,2.5) (nB) {\myfontnow $\featvec_{\nodeind_{\scalebox{.8}1}}$};
   \node at (2,-0.5) (nC) {\myfontnow $\featvec_{\nodeind_{\scalebox{.8}1}}$};
    \node at (1.5,3) (nA) {\myfontnow $\featvec_{\nodeind_{\scalebox{.8}2}}$};
   \node at (5.5,3) (nB) {\myfontnow $\featvec_{\nodeind_{\scalebox{.8}2}}$};
   \node at (3.5,0) (nC) {\myfontnow $\featvec_{\nodeind_{\scalebox{.8}2}}$};
    \node at (0.6,4.2) (nA) {\myfontnow $\featvec_{\nodeind_{\scalebox{.8}3}}$};
   \node at (4.6,4.2) (nB) {\myfontnow $\featvec_{\nodeind_{\scalebox{.8}3}}$};
   \node at (2.6,1.2) (nC) {\myfontnow $\featvec_{\nodeind_{\scalebox{.8}3}}$};
    \node at (2,4) (nA) {\myfontnow $\featvec_{\nodeind_{\scalebox{.8}4}}$};
   \node at (6,4) (nB) {\myfontnow $\featvec_{\nodeind_{\scalebox{.8}4}}$};
   \node at (4,1) (nC) {\myfontnow $\featvec_{\nodeind_{\scalebox{.8}4}}$};
    \node at (2,3) (nA) {\myfontnow $\featvec_{\nodeind_{\scalebox{.8}5}}$};
   \node at (6,3) (nB) {\myfontnow $\featvec_{\nodeind_{\scalebox{.8}5}}$};
   \node at (4,0) (nC) {\myfontnow $\featvec_{\nodeind_{\scalebox{.8}5}}$};
    \node at (0,4) (nA) {\myfontnow $\featvec_{\nodeind_{\scalebox{.8}6}}$};
   \node at (4,4) (nB) {\myfontnow $\featvec_{\nodeind_{\scalebox{.8}6}}$};
   \node at (2,1) (nC) {\myfontnow $\featvec_{\nodeind_{\scalebox{.8}6}}$};
    \node at (-0.5,3.6) (nA) {\myfontnow $\featvec_{\nodeind_{\scalebox{.8}7}}$};
   \node at (3.5,3.6) (nB) {\myfontnow $\featvec_{\nodeind_{\scalebox{.8}7}}$};
   \node at (1.5,0.6) (nC) {\myfontnow $\featvec_{\nodeind_{\scalebox{.8}7}}$};
    \node at (-0.5,3) (nA) {\myfontnow $\featvec_{\nodeind_{\scalebox{.8}8}}$};
   \node at (3.5,3) (nB) {\myfontnow $\featvec_{\nodeind_{\scalebox{.8}8}}$};
   \node at (1.5,0) (nC) {\myfontnow $\featvec_{\nodeind_{\scalebox{.8}8}}$};
   \node at (-0.5,-2) (J){};
  \node at (6.5,6) (L){};
  %\draw[red,thick,dotted] 
 \draw [fill=orange,fill opacity=0.1,draw=none,rounded corners=0.5cm]($(J.north west)+(-0.3,0.6)$) rectangle ($(L.south east)+(0.3,-0.6)$);
    %\draw[thick,dotted]     ($(I.north west)+(-0.5,0.15)$) rectangle ($(L.south east)+(0.5,-0.15)$);

\end{tikzpicture}

%% file: figs/grnn.tex
\begin{tikzpicture}[%
scale=0.9, every node/.style={scale=0.9},
 block/.style={rectangle, fill=gray, text centered, minimum height=8mm, rounded corners=1mm}
        ,my line style1/.style={ draw=purple,thick}
        ,my line style2/.style={ draw=blue,thick}
        ,my line style3/.style={ draw=black,thick}
        ,every label/.append style = {label distance=1pt, inner sep=1pt, 
                             align=left, font=\tiny}
        %,every edge/.style={my line style}
        ,my edge/.style={my line style, mylabel=#1}
        ,my curved edge/.style n args={4}{my line style, out=#1, in=#2, looseness=#3, mylabel=#4}
        ,edge1/.style={my line style1,dashed}
        ,mylabel1/.style={my label style{#1}}
        ,edge2/.style={my line style2, densely dotted}
        ,edge3/.style={my line style3}
        ,mylabel/.style={edge node={node[my label style]{#1}}}]
    \node[nnblock] at (3,0) (nA) {};
    \node[nam,label={[label distance=-0.3cm,text depth=2ex,rotate=90]right:\tiny{\color{orange}\textbf{NAM}}}] at (2.7,0) (namA){};
    \node[gam,label={[label distance=-0.3cm,text depth=2ex,rotate=90]right:\tiny{\color{blue}\textbf{GAM}}}] at (3,0) (gamA){};
    \node[fam,label={[label distance=-0.3cm,text depth=2ex,rotate=90]right:\tiny{\color{green}\textbf{FAM}}}] at (3.3,0) (famA){};
    \node[nnblock] at (3+\distbetweenlayers,0) (nB) {};
        \node[nam,label={[label distance=-0.3cm,text depth=2ex,rotate=90]right:\tiny{\color{orange}\textbf{NAM}}}] at (2.7+\distbetweenlayers,0) (namA){};
    \node[gam,label={[label distance=-0.3cm,text depth=2ex,rotate=90]right:\tiny{\color{blue}\textbf{GAM}}}] at (3+\distbetweenlayers,0) (gamA){};
    \node[fam,label={[label distance=-0.3cm,text depth=2ex,rotate=90]right:\tiny{\color{green}\textbf{FAM}}}] at (3.3+\distbetweenlayers,0) (famA){};
    \node[nnblock] at (3+2*\distbetweenlayers,0) (nC) {};
        \node[nam,label={[label distance=-0.3cm,text depth=2ex,rotate=90]right:\tiny{\color{orange}\textbf{NAM}}}] at (2.7+2*\distbetweenlayers,0) (namA){};
    \node[gam,label={[label distance=-0.3cm,text depth=2ex,rotate=90]right:\tiny{\color{blue}\textbf{GAM}}}] at (3+2*\distbetweenlayers,0) (gamA){};
    \node[fam,label={[label distance=-0.3cm,text depth=2ex,rotate=90]right:\tiny{\color{green}\textbf{FAM}}}] at (3.3+2*\distbetweenlayers,0) (famA){};
    \node[] at (3+3*\distbetweenlayers,0) (ndots) {\huge{$\cdots$}};
    \node[nnblock] at (3+4*\distbetweenlayers,0) (nD) {};
        \node[nam,label={[label distance=-0.3cm,text depth=2ex,rotate=90]right:\tiny{\color{orange}\textbf{NAM}}}] at (2.7+4*\distbetweenlayers,0) (namA){};
    \node[gam,label={[label distance=-0.3cm,text depth=2ex,rotate=90]right:\tiny{\color{blue}\textbf{GAM}}}] at (3+4*\distbetweenlayers,0) (gamA){};
    \node[fam,label={[label distance=-0.3cm,text depth=2ex,rotate=90]right:\tiny{\color{green}\textbf{FAM}}}] at (3.3+4*\distbetweenlayers,0) (famA){};
    \node[outblock,label={[label distance=-0.6cm,text depth=2.8ex,rotate=90]right:\tiny{\textbf{Output layer}}}] at (3+5*\distbetweenlayers,0) (nE) {};
    \definecolor{paramcolor}{HTML}{50ae55}
\definecolor{inputcolor}{HTML}{fd9727}
\definecolor{outputcolor}{HTML}{f1453d}
\tikzset{
	inputmat/.style = {rectangle, draw=inputcolor!70, fill=inputcolor!40, thick, minimum width=1.4cm, minimum height = 1cm},
	outputmat/.style = {rectangle, draw=outputcolor!0, fill=outputcolor!50, thick, minimum width=0.6cm, minimum height = 1.2cm},
	thickermat/.style = {rectangle, draw=black!50, fill=black!20, thick, minimum width=1.4cm, minimum height = 1cm},
	parameter/.style = {rectangle, draw=paramcolor!70, fill=paramcolor!40, thick, minimum width=0.6cm, minimum height = 1.4cm},
}
    \node [outputmat,label={[label distance=-0.5cm,text depth=0ex]right:\footnotesize$\predictionmat$}] at (3+5*\distbetweenlayers+1,0) (nF) {};
    
    \draw [line width=0.1mm] ($(nF.north west) + (0,0.1) $ ) -- ($(nF.north east) + (0,0.1)$); % horizontal line
    \draw [line width=0.1mm] ($(nF.north west) + (0,0.07) $ ) -- ($(nF.north west) + (0,0.13)$); % left tick
\draw [line width=0.1mm] ($(nF.north east) + (0,0.07) $ ) -- ($(nF.north east) + (0,0.13)$); % bottom tick
\node [above=-0.05cm of nF]()[]{\footnotesize $^\nbrclasses$}; % top label

\draw [line width=0.1mm] ($(nF.north east) + (0.1,0) $ ) -- ($(nF.south east) + (0.1,0.0)$); % vertical line
\draw [line width=0.1mm] ($(nF.north east) + (0.13,0) $ ) -- ($(nF.north east) + (0.07,0)$); % top tick
\draw [line width=0.1mm] ($(nF.south east) + (0.13,0) $ ) -- ($(nF.south east) + (0.07,0)$); % bottom tick
\node [right=0cm of nF]()[]{\footnotesize $^{\nbrnodes}$}; % left label

   \node[]  at (-1.5,0) (A) {\begin{turn}{90}
 \input{figs/sampledGraphPlain.tex}\end{turn}   
};       
\draw [decorate,decoration={brace,amplitude=20pt,mirror,raise=4pt},yshift=0pt]
(1,-2) -- (1,2) node [black,midway,xshift=1.8cm]{};

    \newcommand{\xposrel}{-1cm}
    \newcommand{\yposrel}{0.3cm}
    \node[above right = \yposrel and  \xposrel of nA] (eA) {\footnotesize $l=1$};
    \node[above right = \yposrel and  \xposrel of nB] (eB) {\footnotesize $l=2$};
    \node[above right = \yposrel and  \xposrel of nC] (eC) {\footnotesize $l=3$};
    \node[above right = \yposrel and  \xposrel of nD] (eD) {\footnotesize $l=\nbrlayers$};

    \node at (1.3,-2.5) (inp){$\datamatrix$};
    
    \draw[double,->] (nA.east) -- node [above] {\footnotesize $\outputlaytensor\layernot{1}$} (nB);
    \draw[double,->] (nB.east) -- node [above] {\footnotesize$\outputlaytensor\layernot{2}$} (nC);
    \draw[double,->] (nC.east) -- node [above] {\footnotesize$\outputlaytensor\layernot{3}$} (ndots);
    \draw[double,->] (ndots.east) --  (nD);
    \draw[double,->] (nD.east) -- node [above] {\footnotesize$\outputlaytensor\layernot{\nbrlayers}$}(nE);
    \draw[double,->] (nE.east) -- node [above] {}(nF);
    \draw[->, to path={-| (\tikztotarget)}]
        (inp) edge% bend the arrows [bend left]
        (nA) (inp) edge (nB) 
        (inp) edge (nC)(inp) edge (nD);

\end{tikzpicture}

%% file: figs/sampledGraphPlain.tex
\begin{tikzpicture}[%
        scale=0.55, every node/.style={scale=0.55},
        block/.style={rectangle, fill=gray, text centered, minimum height=8mm, rounded corners=1mm}
        ,my line style1/.style={ draw=red,thick}
        ,my line style2/.style={ draw=blue,thick}
        ,my line style3/.style={ draw=black,thick}
        ,every label/.append style = {label distance=1pt, inner sep=1pt, 
                             align=left, font=\tiny}
        %,every edge/.style={my line style}
        ,my edge/.style={my line style, mylabel=#1}
        ,my curved edge/.style n args={4}{my line style, out=#1, in=#2, looseness=#3, mylabel=#4}
        ,edge1/.style={my line style1,dashed}
        ,mylabel1/.style={my label style{#1}}
        ,edge2/.style={my line style2, densely dotted}
        ,edge3/.style={my line style3}
        ,mylabel/.style={edge node={node[my label style]{#1}}}]
  % Place super peers and connect them
\newcommand{\disttopgraphx}{-2}
\newcommand{\disttopgraphy}{+2}
\foreach \place/\name in {{(5+\disttopgraphx,-0.5+\disttopgraphy)/ar0}, {(5.5
+\disttopgraphx,0.3+\disttopgraphy)/br0}, {(5+\disttopgraphx,0.7+\disttopgraphy)/cr0}, {(4.5+\disttopgraphx,0.7+\disttopgraphy)/dr0}}
    \node[superpeers1] (\name) at \place {};
\foreach\place/\name  in {{(6+\disttopgraphx,-0.5+\disttopgraphy)/ad0}, {(6.5+\disttopgraphx,0+\disttopgraphy)/bd0}, {(6+\disttopgraphx,0.7+\disttopgraphy)/cd0}, {(7+\disttopgraphx,0.7+\disttopgraphy)/dd0}}
\node[peers1] (\name) at \place {};

\foreach \distxbetweengraphs\distybetweengraphs\graphid in {{0/-3/1},{3.5/-3/2},{7/-3/3}}
  \foreach \place/\name in {{(0+\distxbetweengraphs,-0.5+\distybetweengraphs)/ar\graphid}, {(0.5+\distxbetweengraphs,0.3+\distybetweengraphs)/br\graphid}, {(0+\distxbetweengraphs,0.7+\distybetweengraphs)/cr\graphid}, {(-0.5+\distxbetweengraphs,0.7+\distybetweengraphs)/dr\graphid}}
    \node[superpeers1] (\name) at \place {};
  
   %
   % Place normal peers
  \foreach \distxbetweengraphs\distybetweengraphs\graphid in  {{0/-3/1},{3.5/-3/2},{7/-3/3}}
  \foreach\place/\name  in {{(1+\distxbetweengraphs,-0.5+\distybetweengraphs)/ad\graphid}, {(1.5+\distxbetweengraphs,0+\distybetweengraphs)/bd\graphid}, {(1+\distxbetweengraphs,0.7+\distybetweengraphs)/cd\graphid}, {(2+\distxbetweengraphs,0.7+\distybetweengraphs)/dd\graphid}}
    \node[peers1] (\name) at \place {};
   \foreach \source/\dest in {ar0/bd0, ar0/cr0, ad0/dd0,bd0/cd0,  br0/cr0, cr0/dr0,
   cd0/dd0,dr0/ad0}
    \path (\source) edge[edge3] (\dest);
    \path (ar0) edge[edge1] (ad0);
    \path (br0) edge[edge1] (bd0);

     \foreach \source/\dest in {ar1/bd1, ad1/dd1,bd1/cd1,  br1/cr1, cr1/dr1,
   cd1/dd1,dr1/ad1}
    \path (\source) edge[edge3] (\dest);
    \path (br1) edge[edge1] (bd1);

  \foreach \source/\dest in {ar2/bd2, ar2/cr2, cr2/dr2,ad2/dd2,br2/cr2,
   cd2/dd2,dr2/ad2}
    \path (\source) edge[edge3] (\dest);
    %\path (ar2) edge[edge1] (ad2);
    
    \foreach \source/\dest in {ar3/bd3, ar3/cr3, ad3/dd3, cr3/dr3,
   cd3/dd3,dr3/ad3}
    \path (\source) edge[edge3] (\dest);
    \path (ar3) edge[edge1] (ad3);
    \path (br3) edge[edge1] (bd3);
    \node[esl] at (3.5,0) (esl1) {\normalsize ED$(q_1,q_2)$};
  %labels 
  %\node[label={[font=\normalsize]0: Corrupted graph $\obsGraph$}] at (4.3+\disttopgraphx,1.5+\disttopgraphy) (cg) {};
   \node[] at (5.5+\disttopgraphx,-0.5+\disttopgraphy) (cg0) {};
  \node[] at (0.5,-2.3) (cg1) {};
  \node[] at (4.2,-2.3) (cg2) {};
  \node[]at (7.8,-2.3) (cg4) {};
  \node[label={[font=\normalsize]0: \large{$\cdots$}}] at (4.2,-1.5) (cg6) {};
  \node[label={[font=\normalsize]0: \Huge{$\cdots$}}] at (5.5,-3) (cg3) {};
  \draw[double,->] (cg0) --  (esl1);
  \draw[dashed,->] (esl1) -- node [above] {$\graph_1$} (cg1);
   \draw[dashed,->] (esl1) -- node [left] {$\graph_2$} (cg2);
    \draw[dashed,->] (esl1) -- node [above] {$\graph_\nbrshifts$} (cg4);

\end{tikzpicture}

%% file: figs/ionosphereaccvsfeatsnr.tex
% This file was created by matlab2tikz.
%
%The latest updates can be retrieved from
%  http://www.mathworks.com/matlabcentral/fileexchange/22022-matlab2tikz-matlab2tikz
%where you can also make suggestions and rate matlab2tikz.
%
\begin{tikzpicture}

\begin{axis}[%
width=0.956\mywidth,
height=0.987\myheight,
at={(0\mywidth,0\myheight)},
scale only axis,
xlabel style={font=\color{white!15!black}},
xlabel={SNR of $\featperturbation$},
xmin=0.2,
xmax=25,
xmode=log,xtick={0.2,1,5,25,125},xticklabels={0.2,1,5,25,125},
ylabel style={font=\color{white!15!black}},
legend columns=4,
xmajorgrids,
ymajorgrids,label style={font=\tiny},
  cycle list name=colorlist,
grid style={dotted},ticklabel style={font=\tiny},
legend style={
	at={(-0.1,1.115)}, 
	anchor=south west, legend cell align=left, align=left,
	draw=none
	% white!15!black
	, font=\small}
]
\addplot%[color=mycolor1, line width=2.0pt, mark=o, mark options={solid, mycolor1}]
  table[row sep=crcr]{%
0.2	0.6\\
1	0.64\\
5 0.75\\
25	0.84\\
};
\addlegendentry{$\kappa=$2}
\addplot
  table[row sep=crcr]{%
0.2	0.74\\
1	0.76\\
5 0.82\\
25	0.83\\
};
\addlegendentry{$\kappa=$5}

\addplot%[color=mycolor2, line width=2.0pt, dashed, mark=square, mark options={solid, mycolor2}]
  table[row sep=crcr]{%
0.2	0.76\\
1	0.76\\
5 0.78\\
25	0.81\\
};
\addlegendentry{$\kappa=$10}
\addplot
%[color=mycolor2, line width=2.0pt, dashed, mark=square, mark options={solid, mycolor2}]
  table[row sep=crcr]{%
0.2	0.78\\
1	0.80\\
5 0.83\\
25	0.85\\
};
\addlegendentry{$\kappa=$5,10}
\end{axis}
\end{tikzpicture}%

%% file: figs/ionosphereaccvsadjsnr.tex
% This file was created by matlab2tikz.
%
%The latest updates can be retrieved from
%  http://www.mathworks.com/matlabcentral/fileexchange/22022-matlab2tikz-matlab2tikz
%where you can also make suggestions and rate matlab2tikz.
%
\begin{tikzpicture}

\begin{axis}[%
width=0.956\mywidth,
height=0.987\myheight,
at={(0\mywidth,0\myheight)},
scale only axis,
xlabel style={font=\xlabelfontsize},
xlabel={SNR of $\shiftperturbation$},
xmin=0.199,
xmax=25.001,
xmode=log,
xtick={0.2,1,5,25,125},
xticklabels={0.2,1,5,25,125},label style={font=\tiny},
ylabel style={font=\xlabelfontsize},
legend columns=4,
xmajorgrids,
ymajorgrids,
  cycle list name=colorlist,
grid style={dotted},ticklabel style={font=\tiny},
legend style={
	at={(0,-0.415)}, 
	anchor=south west, legend cell align=left, align=left,
	draw=none
	% white!15!black
	, font=\legendfontsize}
]
\addplot%[color=mycolor1, line width=2.0pt, mark=o, mark options={solid, mycolor1}]
  table[row sep=crcr]{%
0.2	0.58\\
1	0.61\\
5 0.65\\
25	0.75\\
125	0.84\\
};
%\addlegendentry{$\kappa=$2}
\addplot
  table[row sep=crcr]{%
0.2	0.59\\
1	0.58\\
5 0.69\\
25	0.72\\
125	0.84\\
};
%\addlegendentry{$\kappa=$5}
\addplot%[color=mycolor2, line width=2.0pt, dashed, mark=square, mark options={solid, mycolor2}]
  table[row sep=crcr]{%
0.2	0.60\\
1	0.62\\
5 0.63\\
25	0.74\\
125	0.85\\
};
%\addlegendentry{$\kappa=$10}
\addplot
%[color=mycolor2, line width=2.0pt, dashed, mark=square, mark options={solid, mycolor2}]
  table[row sep=crcr]{%
0.2	0.63\\
1	0.67\\
5 0.75\\
25	0.80\\
125	0.88\\
};
\end{axis}
\end{tikzpicture}%

%% file: figs/adlinkscora.tex
% This file was created by matplotlib2tikz v0.6.18.
\begin{tikzpicture}

\definecolor{color0}{rgb}{1,0.647058823529412,0}
\definecolor{color1}{rgb}{1,1,0}
\definecolor{color2}{rgb}{0.501960784313725,0,0.501960784313725}

\begin{axis}[width=0.956\mywidths,
height=0.987\myheights,
at={(0\mywidth,0\myheight)},
%legend entries={{custom-nonlin-invariant},{custom-lin-invariant},{custom-nonlin-raw},{diff-nonlin-invariant},{diff-nonlin-raw},{monicCubic-nonlin-invariant},{monicCubic-nonlin-raw},
%{pointnet}},
legend style={draw=white!80.0!black},
tick align=outside,
tick pos=left,scale only axis,
x grid style={white!69.01960784313725!black},
xmin=1, xmax=100000, xmode=log,
xtick={1,100,10000},
%xticklabel={\pgfmathparse{\tick*100/10312}\pgfmathprintnumber{\pgfmathresult}\%},
%y grid style={white!69.01960784313725!black},
%ylabel={accuracy},
%ymin=0.356170284628286, ymax=0.858165821954202,
xmajorgrids,
ymajorgrids,ticklabel style={font=\tiny},
grid style={dotted},
legend columns=2,
legend style={
	at={(0,1.015)}, 
	anchor=south west, legend cell align=left, align=left, draw=none
	% white!15!black
	, font=\legendfontsize}]

\addplot [line width=\mylinewidth,AGNN, mark=*, mark size=\markwidth, mark options={solid}]
table [row sep=\\]{%
1 0.685\\
1000 0.6820003390312195\\%'max,config:features=0,model=agrcn_edgesamplelearn_rate=0.05,smooth_reg=1e-06,hidden_units1=32hidden_units2=0epochs=250,dropout_rate=0.1,weight_decay=1e-06early_stopping=20,neighbor_list=[],sampl_nbr=2,sample_pct=0.99,max_degree=3,sparse_reg=1e-06': 0.6820003390312195}
10000   0.6490002512931824\\		%config:features=0,model=agrcn_edgesamplelearn_rate=0.05,smooth_reg=1e-06,hidden_units1=32hidden_units2=0epochs=250,dropout_rate=0.1,weight_decay=1e-06early_stopping=20,neighbor_list=[],sampl_nbr=2,sample_pct=0.98,max_degree=3,sparse_reg=1e-06': 0.6490002512931824}
50000 0.636000764369964\\%'max,config:features=0,model=agrcn_edgesamplelearn_rate=0.05,smooth_reg=1e-06,hidden_units1=32hidden_units2=0epochs=250,dropout_rate=0.1,weight_decay=1e-06early_stopping=20,neighbor_list=[],sampl_nbr=2,sample_pct=0.99,max_degree=3,sparse_reg=1e-06': 0.5936000764369964
};%\addlegendentry{AGCN}
\addplot [mark size=\markwidth, line width=\mylinewidth, GCN, dashed, mark=x,  mark options={solid}]
table [row sep=\\]{%
1 0.6784003496170044\\%'max,config:features=0,model=gcnlearn_rate=0.05,smooth_reg=1e-06,hidden_units1=64hidden_units2=0epochs=250,dropout_rate=0.1,weight_decay=1e-06early_stopping=100,neighbor_list=[],sampl_nbr=2,sample_pct=0.98,max_degree=3,sparse_reg=1e-06': 0.6784003496170044
1000 0.6460002422332763\\%ax,config:features=0,model=gcnlearn_rate=0.05,smooth_reg=1e-06,hidden_units1=64hidden_units2=8epochs=250,dropout_rate=0.7,weight_decay=1e-06early_stopping=20,neighbor_list=[],sampl_nbr=2,sample_pct=0.98,max_degree=3,sparse_reg=1e-06': 0.6460002422332763}
10000  0.633200216293335			\\%,config:features=0,model=gcnlearn_rate=0.05,smooth_reg=0.01,hidden_units1=32hidden_units2=8epochs=250,dropout_rate=0.1,weight_decay=1e-06early_stopping=20,neighbor_list=[],sampl_nbr=2,sample_pct=0.98,max_degree=3,sparse_reg=1e-06': 0.633200216293335
50000 0.5389999330043793\\%'max,config:features=0,model=gcnlearn_rate=0.05,smooth_reg=1e-06,hidden_units1=32hidden_units2=0epochs=250,dropout_rate=0.7,weight_decay=1e-06early_stopping=20,neighbor_list=[],sampl_nbr=2,sample_pct=0.98,max_degree=3,sparse_reg=1e-06': 0.5389999330043793}
};
%\addlegendentry{GCN}
\end{axis}
\end{tikzpicture}

%% file: figs/adlinkspubmed.tex
% This file was created by matplotlib2tikz v0.6.18.
\begin{tikzpicture}

\begin{axis}[width=0.956\mywidths,
height=0.987\myheights,
at={(0\mywidth,0\myheight)},
legend style={draw=white!80.0!black},
tick align=outside,scale only axis,
tick pos=left,
x grid style={white!69.01960784313725!black},
xmin=1, xmax=100000, xmode=log,
xtick={1,100,10000},%xticklabel={\pgfmathparse{\tick*100/10312}\pgfmathprintnumber{\pgfmathresult}\%},
%y grid style={white!69.01960784313725!black},
%ylabel={accuracy},
%ymin=0.356170284628286, ymax=0.858165821954202,
xmajorgrids,
ymajorgrids,
grid style={dotted},
legend columns=2,ticklabel style={font=\tiny},
legend style={
	at={(0.4,1.015)}, 
	anchor=south west, legend cell align=left, align=left, draw=none
	% white!15!black
	, font=\small}]

\addplot [line width=\mylinewidth,AGNN, mark=*, mark size=\markwidth, mark options={solid}]
table [row sep=\\]{%
1 0.668008856773377\\
1000   0.6570007920265197\\		%'max,config:features=0,model=gcnlearn_rate=0.05,smooth_reg=1e-06,hidden_units1=32hidden_units2=8epochs=250,dropout_rate=0.1,weight_decay=1e-06early_stopping=20,neighbor_list=[],sampl_nbr=2,sample_pct=0.98,max_degree=3,sparse_reg=1e-06': 0.6570007920265197}
10000 0.6494006896018982\\%'max,config:features=0,model=gcnlearn_rate=0.05,smooth_reg=1e-06,hidden_units1=64hidden_units2=0epochs=250,dropout_rate=0.1,weight_decay=1e-06early_stopping=20,neighbor_list=[],sampl_nbr=2,sample_pct=0.98,max_degree=3,sparse_reg=1e-06': 0.6494006896018982}
50000 0.6354006671905517\\%'max,config:features=0,model=gcnlearn_rate=0.05,smooth_reg=1e-06,hidden_units1=64hidden_units2=0epochs=250,dropout_rate=0.1,weight_decay=1e-06early_stopping=20,neighbor_list=[],sampl_nbr=2,sample_pct=0.98,max_degree=3,sparse_reg=1e-06': 0.6154006671905517}
};\addlegendentry{AGCN}

\addplot [mark size=\markwidth, line width=\mylinewidth, GCN, dashed, mark=x,  mark options={solid}]
table [row sep=\\]{%\
1 0.6580008029937744\\
1000 0.6426008152961731\\%'max,config:features=0,model=gcnlearn_rate=0.05,smooth_reg=0.0001,hidden_units1=32hidden_units2=0epochs=250,dropout_rate=0.1,weight_decay=1e-06early_stopping=20,neighbor_list=[],sampl_nbr=10,sample_pct=0.99,max_degree=3,sparse_reg=0.0001': 0.6426008152961731}
10000  0.6364007735252381			\\%'max,config:features=0,model=gcnlearn_rate=0.05,smooth_reg=1e-06,hidden_units1=32hidden_units2=0epochs=250,dropout_rate=0.1,weight_decay=1e-06early_stopping=20,neighbor_list=[],sampl_nbr=2,sample_pct=0.98,max_degree=3,sparse_reg=0.0001': 0.6364007735252381}
50000 0.6000762462616\\%'max,config:features=0,model=gcnlearn_rate=0.05,smooth_reg=0.0001,hidden_units1=64hidden_units2=0epochs=250,dropout_rate=0.1,weight_decay=1e-06early_stopping=20,neighbor_list=[],sampl_nbr=5,sample_pct=0.98,max_degree=3,sparse_reg=0.01': 0.618000762462616}
};\addlegendentry{GCN}

\end{axis}
\end{tikzpicture}

%% file: figs/adlinkciteseer.tex
% This file was created by matplotlib2tikz v0.6.18.
\begin{tikzpicture}

\definecolor{color0}{rgb}{1,0.647058823529412,0}
\definecolor{color1}{rgb}{1,1,0}
\definecolor{color2}{rgb}{0.501960784313725,0,0.501960784313725}

\begin{axis}[width=0.956\mywidths,
height=0.987\myheights,
at={(0\mywidth,0\myheight)},
legend style={draw=white!80.0!black},
tick align=outside,
tick pos=left,scale only axis,
x grid style={white!69.01960784313725!black},
xmin=1, xmax=100000, xmode=log,ticklabel style={font=\tiny},
xtick={1,100,10000},%xticklabel={\pgfmathparse{\tick*100/10312}\pgfmathprintnumber{\pgfmathresult}\%},
%y grid style={white!69.01960784313725!black},
%ylabel={accuracy},
%ymin=0.356170284628286, ymax=0.858165821954202,
xmajorgrids,
ymajorgrids,
grid style={dotted},
legend columns=2,
legend style={
	at={(0,1.015)}, 
	anchor=south west, legend cell align=left, align=left, draw=none
	% white!15!black
	, font=\legendfontsize}]

\addplot [line width=\mylinewidth,AGNN, mark=*, mark size=\markwidth, mark options={solid}]
table [row sep=\\]{%
1 0.479973134994507\\
1000   0.4695997323989868\\		%'max,config:features=0,model=gcnlearn_rate=0.05,smooth_reg=0.0001,hidden_units1=32hidden_units2=0epochs=250,dropout_rate=0.1,weight_decay=1e-06early_stopping=20,neighbor_list=[],sampl_nbr=5,sample_pct=0.98,max_degree=3,sparse_reg=1e-06': 0.4695997323989868}
10000 0.467799711227417\\%'max,config:features=0,model=gcnlearn_rate=0.05,smooth_reg=0.01,hidden_units1=64hidden_units2=0epochs=250,dropout_rate=0.7,weight_decay=1e-06early_stopping=20,neighbor_list=[],sampl_nbr=2,sample_pct=0.98,max_degree=3,sparse_reg=0.0001': 0.467799711227417}
50000 0.44719977498054507\\%'max,config:features=0,model=gcnlearn_rate=0.05,smooth_reg=0.0001,hidden_units1=32hidden_units2=8epochs=250,dropout_rate=0.1,weight_decay=1e-06early_stopping=20,neighbor_list=[],sampl_nbr=10,sample_pct=0.99,max_degree=3,sparse_reg=1e-06': 0.42719977498054507}
};%\addlegendentry{AGCN}

\addplot [mark size=\markwidth, line width=\mylinewidth, GCN, dashed, mark=x,  mark options={solid}]
table [row sep=\\]{%
1 0.461599725484848\\
1000 0.4563997304439545\\%'max,config:features=0,model=gcnlearn_rate=0.05,smooth_reg=1e-06,hidden_units1=32hidden_units2=8epochs=250,dropout_rate=0.1,weight_decay=1e-06early_stopping=20,neighbor_list=[],sampl_nbr=2,sample_pct=0.98,max_degree=3,sparse_reg=1e-06': 0.4563997304439545}
10000  0.4425997519493103			\\%'max,config:features=0,model=gcnlearn_rate=0.05,smooth_reg=1e-06,hidden_units1=32hidden_units2=0epochs=250,dropout_rate=0.7,weight_decay=1e-06early_stopping=20,neighbor_list=[],sampl_nbr=2,sample_pct=0.98,max_degree=3,sparse_reg=1e-06': 0.4425997519493103}
50000 0.4109997868537903\\%'max,config:features=0,model=gcnlearn_rate=0.05,smooth_reg=1e-06,hidden_units1=32hidden_units2=0epochs=250,dropout_rate=0.7,weight_decay=1e-06early_stopping=20,neighbor_list=[],sampl_nbr=2,sample_pct=0.98,max_degree=3,sparse_reg=1e-06': 0.4109997868537903}
};%\addlegendentry{GCN}

\end{axis}
\end{tikzpicture}

%% file: figs/adlinkspolblog.tex
% This file was created by matplotlib2tikz v0.6.18.
\begin{tikzpicture}

\definecolor{color0}{rgb}{1,0.647058823529412,0}
\definecolor{color1}{rgb}{1,1,0}
\definecolor{color2}{rgb}{0.501960784313725,0,0.501960784313725}

\begin{axis}[width=0.956\mywidths,
height=0.987\myheights,
at={(0\mywidth,0\myheight)},
%legend entries={{custom-nonlin-invariant},{custom-lin-invariant},{custom-nonlin-raw},{diff-nonlin-invariant},{diff-nonlin-raw},{monicCubic-nonlin-invariant},{monicCubic-nonlin-raw},
%{pointnet}},
legend style={draw=white!80.0!black},
tick align=outside,
tick pos=left,scale only axis,
x grid style={white!69.01960784313725!black},
xmin=1, xmax=100000, xmode=log,
xtick={1,100,10000},
%xticklabel={\pgfmathparse{\tick*100/10312}\pgfmathprintnumber{\pgfmathresult}\%},
%y grid style={white!69.01960784313725!black},
%ylabel={accuracy},
%ymin=0.356170284628286, ymax=0.858165821954202,
xmajorgrids,
ymajorgrids,
grid style={dotted},
legend columns=2,ticklabel style={font=\tiny},
legend style={
	at={(0,1.015)}, 
	anchor=south west, legend cell align=left, align=left, draw=none
	% white!15!black
	, font=\legendfontsize}]

\addplot [line width=\mylinewidth,AGNN, mark=*, mark size=\markwidth, mark options={solid}]
table [row sep=\\]{%
1 0.9602509498596191\\
1000 0.9414224982261657\\%'max,config:features=0,model=agrcn_edgesamplelearn_rate=0.05,smooth_reg=1e-06,hidden_units1=32hidden_units2=0epochs=250,dropout_rate=0.1,weight_decay=1e-06early_stopping=20,neighbor_list=[],sampl_nbr=2,sample_pct=0.99,max_degree=3,sparse_reg=1e-06': 0.6820003390312195}
10000   0.9336819410324096\\		%config:features=0,model=agrcn_edgesamplepqlearn_rate=0.05,smooth_reg=1e-08,hidden_units1=64hidden_units2=8epochs=250,dropout_rate=0.7,weight_decay=0.1early_stopping=80,neighbor_list=[],sampl_nbr=10,sample_pct=0.0,p=0.99,q=0.0,max_degree=3,sparse_reg=1e-08:=0.9414224982261657
50000 0.8215480446815491\\%'config:features=0,model=agrcn_edgesamplepqlearn_rate=0.05,smooth_reg=1e-08,hidden_units1=64hidden_units2=8epochs=250,dropout_rate=0.7,weight_decay=1e-08early_stopping=80,neighbor_list=[],sampl_nbr=10,sample_pct=0.0,p=0.99,q=0.0,max_degree=3,sparse_reg=1e-08:=0.8215480446815491
};%\addlegendentry{AGCN}
\addplot [mark size=\markwidth, line width=\mylinewidth, GCN, dashed, mark=x,  mark options={solid}]
table [row sep=\\]{%
1 0.950209128856659\\%''max,config:features=0,model=gcnlearn_rate=0.05,smooth_reg=1e-06,hidden_units1=64hidden_units2=8epochs=250,dropout_rate=0.1,weight_decay=0.1early_stopping=100,neighbor_list=[],sampl_nbr=2,sample_pct=0.98,max_degree=3,sparse_reg=1e-06': 0.9383663296699524}
1000 0.9334726691246032\\%ax,config:features=0,model=gcnlearn_rate=0.05,smooth_reg=1e-06,hidden_units1=64hidden_units2=8epochs=250,dropout_rate=0.7,weight_decay=1e-06early_stopping=20,neighbor_list=[],sampl_nbr=2,sample_pct=0.98,max_degree=3,sparse_reg=1e-06': 0.6460002422332763}
10000  0.9260250258445739			\\%,config:features=0,model=gcnlearn_rate=0.05,smooth_reg=0.01,hidden_units1=32hidden_units2=8epochs=250,dropout_rate=0.1,weight_decay=1e-06early_stopping=20,neighbor_list=[],sampl_nbr=2,sample_pct=0.98,max_degree=3,sparse_reg=1e-06': 0.633200216293335
50000 0.5202928304672241\\%'max,config:features=0,model=gcnlearn_rate=0.05,smooth_reg=1e-06,hidden_units1=32hidden_units2=0epochs=250,dropout_rate=0.7,weight_decay=1e-06early_stopping=20,neighbor_list=[],sampl_nbr=2,sample_pct=0.98,max_degree=3,sparse_reg=1e-06': 0.5389999330043793}
};
%\addlegendentry{GCN}
\end{axis}
\end{tikzpicture}

%% file: figs/agcnq2per.tex
% This file was created by matplotlib2tikz v0.6.18.
\begin{tikzpicture}

\begin{axis}[width=0.956\mywidthss,
height=0.987\myheightss,
at={(0\mywidth,0\myheight)},
%legend entries={{custom-nonlin-invariant},{custom-lin-invariant},{custom-nonlin-raw},{diff-nonlin-invariant},{diff-nonlin-raw},{monicCubic-nonlin-invariant},{monicCubic-nonlin-raw},
%{pointnet}},
legend style={draw=white!80.0!black},
tick align=outside,
tick pos=left,scale only axis,
x grid style={white!69.01960784313725!black},
xlabel={$1-q_2$},
xmin=0.00001, xmax=0.1, xmode=log,
xtick={0.0001,0.01},
%xticklabel={\pgfmathparse{\tick*100/10312}\pgfmathprintnumber{\pgfmathresult}\%},
%y grid style={white!69.01960784313725!black},
%ylabel={accuracy},
%ymin=0.356170284628286, ymax=0.858165821954202,
xmajorgrids,
ymajorgrids,ticklabel style={font=\tiny},
grid style={dotted},label style={font=\tiny},
legend columns=1,
ytick={0.7,0.8},
ticklabel style={font=\tiny,  inner sep=0pt,outer sep=0pt},
legend style={
	at={(0,1.015)}, 
	anchor=south west, legend cell align=left, align=left, draw=none
	% white!15!black
	, font=\tiny}]

\addplot [line width=0.5*\mylinewidth,AGNN, mark=*, mark size=\markwidth, mark options={solid}]
table [row sep=\\]{%
0.00001 0.7733333826065063\\
0.0001 0.7866667151451111\\%'max,config:features=0,model=agrcn_edgesamplelearn_rate=0.05,smooth_reg=1e-06,hidden_units1=32hidden_units2=0epochs=250,dropout_rate=0.1,weight_decay=1e-06early_stopping=20,neighbor_list=[],sampl_nbr=2,sample_pct=0.99,max_degree=3,sparse_reg=1e-06': 0.6820003390312195}
0.001  0.79333336353302\\		%config:features=0,model=agrcn_edgesamplelearn_rate=0.05,smooth_reg=1e-06,hidden_units1=32hidden_units2=0epochs=250,dropout_rate=0.1,weight_decay=1e-06early_stopping=20,neighbor_list=[],sampl_nbr=2,sample_pct=0.98,max_degree=3,sparse_reg=1e-06': 0.6490002512931824}
0.01 0.6733333706855774\\%'max,config:features=0,model=agrcn_edgesamplelearn_rate=0.05,smooth_reg=1e-06,hidden_units1=32hidden_units2=0epochs=250,dropout_rate=0.1,weight_decay=1e-06early_stopping=20,neighbor_list=[],sampl_nbr=2,sample_pct=0.99,max_degree=3,sparse_reg=1e-06': 0.5936000764369964
0.1 0.6666667103767395\\
};
%\addlegendentry{GCN}
\end{axis}
\end{tikzpicture}

%% file: figs/agcnp2per.tex
% This file was created by matplotlib2tikz v0.6.18.
\begin{tikzpicture}

\begin{axis}[width=0.956\mywidthss,
height=0.987\myheightss,
at={(0\mywidth,0\myheight)},
%legend entries={{custom-nonlin-invariant},{custom-lin-invariant},{custom-nonlin-raw},{diff-nonlin-invariant},{diff-nonlin-raw},{monicCubic-nonlin-invariant},{monicCubic-nonlin-raw},
%{pointnet}},
legend style={draw=white!80.0!black},
tick align=outside,
tick pos=left,scale only axis,
x grid style={white!69.01960784313725!black},
xlabel={$q_1$},
xmin=0.8, xmax=1,label style={font=\tiny},
xtick={0.8,0.9,1},
ytick={0.75,0.8},
ticklabel style={font=\tiny, inner sep=0pt,outer sep=0pt},
%xticklabel={\pgfmathparse{\tick*100/10312}\pgfmathprintnumber{\pgfmathresult}\%},
%y grid style={white!69.01960784313725!black},
%ylabel={accuracy},
%ymin=0.356170284628286, ymax=0.858165821954202,
xmajorgrids,
ymajorgrids,ticklabel style={font=\tiny},
grid style={dotted},
legend columns=1,
legend style={
	at={(0,1.015)}, 
	anchor=south west, legend cell align=left, align=left, draw=none
	% white!15!black
	, font=\legendfontsize}]

\addplot [line width=0.5*\mylinewidth,AGNN, mark=*, mark size=\markwidth, mark options={solid}]
table [row sep=\\]{%
0.8 0.7466667103767395\\
0.85 0.7866667079925537\\
0.9  0.8006667079925537\\	
0.99 0.79333336353302\\
};%\addlegendentry{AGCN}
%\addlegendentry{GCN}
\end{axis}
\end{tikzpicture}

%% file: figs/agcnI2per.tex
% This file was created by matplotlib2tikz v0.6.18.
\begin{tikzpicture}

\begin{axis}[width=0.956\mywidthss,
height=0.987\myheightss,
at={(0\mywidth,0\myheight)},
%legend entries={{custom-nonlin-invariant},{custom-lin-invariant},{custom-nonlin-raw},{diff-nonlin-invariant},{diff-nonlin-raw},{monicCubic-nonlin-invariant},{monicCubic-nonlin-raw},
%{pointnet}},
legend style={draw=white!80.0!black},
tick align=outside,
tick pos=left,scale only axis,
x grid style={white!69.01960784313725!black},
xlabel={$I$},
xmin=1, xmax=21,
%xtick={0.0001,0.01},
%xticklabel={\pgfmathparse{\tick*100/10312}\pgfmathprintnumber{\pgfmathresult}\%},
%y grid style={white!69.01960784313725!black},
%ylabel={accuracy},
%ymin=0.356170284628286, ymax=0.858165821954202,
xmajorgrids,
ymajorgrids,ticklabel style={font=\tiny},
grid style={dotted},
ytick={0.75,0.8},
ticklabel style={font=\tiny, inner sep=0pt,outer sep=0pt},
legend columns=1,label style={font=\tiny},
legend style={
	at={(0,1.015)}, 
	anchor=south west, legend cell align=left, align=left, draw=none
	% white!15!black
	, font=\tiny}]

\addplot [line width=0.5*\mylinewidth,AGNN, mark=*, mark size=\markwidth, mark options={solid}]
table [row sep=\\]{%
2 0.7433333945274353\\
5 0.7866666913032532\\
10  0.8000000238418579\\		%config:features=0,model=agrcn_edgesamplelearn_rate=0.05,smooth_reg=1e-06,hidden_units1=32hidden_units2=0epochs=250,dropout_rate=0.1,weight_decay=1e-06early_stopping=20,neighbor_list=[],sampl_nbr=2,sample_pct=0.98,max_degree=3,sparse_reg=1e-06': 0.6490002512931824}
15 0.8002000238418579\\%'max,config:features=0,model=agrcn_edgesamplelearn_rate=0.05,smooth_reg=1e-06,hidden_units1=32hidden_units2=0epochs=250,dropout_rate=0.1,weight_decay=1e-06early_stopping=20,neighbor_list=[],sampl_nbr=2,sample_pct=0.99,max_degree=3,sparse_reg=1e-06': 0.5936000764369964
20 0.8008000238418579\\
};
%\addlegendentry{GCN}
\end{axis}
\end{tikzpicture}